\documentclass{ws-procs11x85}
\usepackage{ws-procs-thm}           
\usepackage{graphicx}
\usepackage{longtable}
\usepackage{color}
\usepackage{multirow}
\usepackage{tabularx}
\usepackage{amsmath}
\usepackage{booktabs}
\usepackage{wrapfig}
\usepackage{titlesec}
\titlespacing*{\section}{0pt}{0.5\baselineskip}{0.5\baselineskip}
\titlespacing*{\subsection}{0pt}{0.5\baselineskip}{0.5\baselineskip}
\titlespacing*{\paragraph}{0pt}{0.5\baselineskip}{0.5\baselineskip}

\begin{document}

\title{Optimization of Genomic Classifiers for Clinical Deployment: Evaluation of Bayesian Optimization to Select Predictive Models of Acute Infection and In-Hospital Mortality\footnote{Supplementary material can be found at \url{https://arxiv.org/abs/2003.12310}}}

\author{Michael B. Mayhew$^\dag$, Elizabeth Tran, Kirindi Choi, Uros Midic, Roland Luethy, Nandita Damaraju and Ljubomir Buturovic}

\address{Inflammatix, Inc.\\
Burlingame, California 94010, USA\\
$^\dag$E-mail: mmayhew@inflammatix.com\\
www.inflammatix.com}

\begin{abstract}
Acute infection, if not rapidly and accurately detected, can lead to sepsis, organ failure and even death. Current detection of acute infection as well as assessment of a patient's severity of illness are imperfect. Characterization of a patient's immune response by quantifying expression levels of specific genes from blood represents a potentially more timely and precise means of accomplishing both tasks. Machine learning methods provide a platform to leverage this \emph{host response} for development of deployment-ready classification models. Prioritization of promising classifiers is dependent, in part, on hyperparameter optimization for which a number of approaches including grid search, random sampling and Bayesian optimization have been shown to be effective. We compare HO approaches for the development of diagnostic classifiers of acute infection and in-hospital mortality from gene expression of 29 diagnostic markers. We take a deployment-centered approach to our comprehensive analysis, accounting for heterogeneity in our multi-study patient cohort with our choices of dataset partitioning and hyperparameter optimization objective as well as assessing selected classifiers in external (as well as internal) validation. We find that classifiers selected by Bayesian optimization for in-hospital mortality can outperform those selected by grid search or random sampling. However, in contrast to previous research: 1) Bayesian optimization is not more efficient in selecting classifiers in all instances compared to grid search or random sampling-based methods and 2) we note marginal gains in classifier performance in only specific circumstances when using a common variant of Bayesian optimization (i.e. automatic relevance determination). Our analysis highlights the need for further practical, deployment-centered benchmarking of HO approaches in the healthcare context. 
\end{abstract}


\copyrightinfo{\copyright\ 2020 The Authors. Open Access chapter published by World Scientific Publishing Company and distributed under the terms of the Creative Commons Attribution Non-Commercial (CC BY-NC) 4.0 License.}

\section{Introduction}
Patient lives depend on the swiftness and accuracy of 1) assessment of the severity of their illness and 2) detection of acute infection (when present). The COVID-19 pandemic has put this fact into stark relief. Currently, clinicians determine severity of illness by computing scores (e.g. SOFA~\cite{Jones2009}) based on patient physiological features associated with the risk of adverse events (e.g. in-hospital mortality, organ failure). Similarly, detection of acute infection generally involves evaluation of symptoms (e.g. cough, runny nose, fever) as well as laboratory tests for the presence of specific pathogens. However, these methods provide superficial and imprecise measures of patient illness. Recent work has highlighted the potential of using gene expression measurements from patient blood to detect the presence and type of infection to which the patient is responding~\cite{sweeney2015comprehensive,sweeney2016robust,sweeney2017benchmarking,Mayhew2020} as well as the patient's severity of illness~\cite{Sweeney2018}. 

Coupled with these host response signatures, advances in machine learning (ML) provide a platform for the development of robust, diagnostic classifiers of acute infection status (e.g. bacterial or viral) and in-hospital mortality from gene expression. An important step in this development is optimization of the classifier's hyperparameters (e.g. penalty coefficient in a LASSO logistic regression, learning rates for gradient descent). Hyperparameter optimization begins with specification of a search space and proceeds by generating a user-specified number of hyperparameter configurations, training the classifier models given by each configuration, and evaluating the performance of the trained classifier in \emph{internal validation}. Internal validation performance is typically assessed either on a separate validation/tuning dataset or by cross-validation. Configurations are then ranked by this performance, with the top configuration selected and retained for \emph{external validation} (application to a held-out dataset). 

Multiple HO approaches have been proposed. For classifiers with relatively small hyperparameter spaces (e.g. support vector machines), optimizing over a pre-defined grid of hyperparameter values (grid search; GS) has proven effective. More recent work has shown that optimization by randomly sampling (RS) hyperparameter configurations can lead to better coverage of high-dimensional hyperparameter spaces and potentially better classifier performance~\cite{bergstra2012random}. Bayesian optimization (BO) is a global optimization procedure that has also proven effective for hyperparameter optimization in classical~\cite{Snoek2012,snoek2015scalable,Klein2017,Falkner2018,Klein2019} and biomedical~\cite{Ghassemi2014,Colopy2018,Nishio2018,Borgli2019} ML applications. In BO, one uses a model (commonly a Gaussian process (GP)~\cite{Rasmussen2006}) to approximate the objective function one wants to optimize; for hyperparameter optimization, the objective function maps from hyperparameter configurations to the internal validation performance of their corresponding classifiers. In contrast to GS/RS, BO proceeds by sequentially evaluating configurations with each newly visited configuration used to update the model of the objective function.

In this work, we compare GS/RS and BO methods for hyperparameter optimization of gene expression-based diagnostic classifiers for two clinical tasks: 1) detection of acute infection and 2) prediction of mortality within 30 days of hospitalization. We optimize and train three different types of classifiers using gene expression features from 29 diagnostic markers in a multi-study cohort of 3413 patient samples for acute infection detection (3288 for 30-day mortality prediction). Patient samples were assayed on a variety of technical platforms and collected from a range of geographical regions, healthcare settings, and disease contexts. Our extensive analysis evaluates the BO approach, in particular, under a range of computational budgets and optimization settings. Crucially, beyond assessing and comparing the performance of top classifiers in internal validation, we further evaluate top models selected by all HO approaches in a multi-cohort external validation set comprising nearly 300 patients profiled by a targeted diagnostic instrument (NanoString). Our analysis provides important insights for diagnostic classifier development using genomic data, and, more generally, about the implementation and practical usage of HO methods in healthcare.

\section{Related Work}
Previous studies comparing HO approaches in the ML community have demonstrated that BO can select promising classifiers more efficiently (with fewer evaluations of hyperparameter configurations) than GS/RS methods~\cite{bergstra2013making,Snoek2012,snoek2015scalable,Klein2017,Falkner2018,Nishio2018,Borgli2019,Klein2019}. However, these studies have focused on internal validation performance and on benchmark datasets whose composition and handling (i.e. partitioning into training-validation-test splits) doesn't necessarily reflect characteristics of healthcare settings (i.e. smaller, structured, and more heterogeneous datasets; high propensity for models to be applied to out-of-distribution samples at test time~\cite{BenDavid2007}). 

Bayesian optimization has also found recent success in genomics and biomedical applications~\cite{Thomas2018,Tanaka2018,Mao2019}. Ghassemi et al.~\cite{Ghassemi2014} compare multiple HO approaches, including BO, for tuning parameters of the multi-scale entropy of heart rate time series to aid mortality prediction among sepsis patients. Colopy et al.~\cite{Colopy2018} analyzed RS and BO methods for optimization of patient-specific GP regression models used in vital-sign forecasting. A study by Nishio et al.~\cite{Nishio2018} evaluated both RBF SVM and XGBoost classifiers tuned by either RS or BO for detection of lung cancer from nodule CT scans. Borgli et al.~\cite{Borgli2019} evaluated BO for tuning and transfer learning of pre-trained convolutional neural networks to detect gastrointestinal conditions from images. Again, however, these studies only reported either internal validation performance or performance on a test set partitioned from a full, relatively small and homogeneous (e.g. collected from a single hospital) dataset, making conclusions difficult to draw about the generalizability of selected models in other segments of the deployment population. Moreover, these studies focused on: 1) no more than two classifier types, 2) a narrow range of settings for BO, and 3) physiological or image data. To our knowledge, no studies have evaluated the external validation performance of selected models, an important pre-requisite for eventual model deployment. In addition, no comparison of HO approaches has yet been attempted for development of diagnostic classifiers using genomic data.

\section{Methods}
\subsection{Cohort \& Feature Description}
To build our datasets, we combined gene expression data from public sources and in-house clinical studies designed for research in diagnosing acute infections and sepsis. We collected the publicly available studies from the NCBI GEO and EMBL-EBI ArrayExpress databases using a systematic search~\cite{sweeney2015comprehensive}. The public studies were profiled using a variety of different technical platforms (e.g. mostly microarrays). Samples from the in-house clinical studies were profiled on the NanoString nCounter platform using a custom codeset for 29 diagnostic genes of interest. All included studies consisted of samples from our target population: both adult and pediatric patients from diverse geographical regions and clinical settings. Each included study had measurements taken from patient blood for all 29 markers. To account for heterogeneity across studies, we performed co-normalization (see~\cite{Mayhew2020} and the Supplement).

The features we used in our analyses were based on the expression values of 29 genes previously found to accurately discriminate three different aspects of acute infection: 1) viral vs. bacterial infection (7 genes)~\cite{sweeney2016robust}, 2) infection vs. non-infectious inflammation (11 genes)~\cite{sweeney2015comprehensive}, and 3) high vs. low risk of 30-day mortality (11 genes)~\cite{Sweeney2018}. Building on our previous work~\cite{Mayhew2020}, we computed both the geometric means and arithmetic means of these six groups of genes, producing 12 features. We optimized and trained our classifiers on the combination of these 12 features and the expression values of all 29 genes (41 features in total). Labels for one of three classes of the acute infection detection or BVN task (\textbf{B}acterial infection, \textbf{V}iral infection, or \textbf{N}on-infectious inflammation) were determined differently for each of the training and validation studies depending on available data. For training set studies, we used the labels provided by each study, deferring to each study's criteria for adjudication which may have involved multi-clinician adjudication with or without positive pathogen identification or positive pathogen identification alone. When BVN adjudications were not directly provided by the study, we assigned class labels based on available pathogen test results from the study metadata/manuscripts. For validation data, one study was adjudicated by a panel of clinicians using all available clinical data (including pathogen test results) while all other validation studies were labeled by us using only pathogen test results. Non-infected determinations did not include healthy controls. Binary indicator labels of whether a patient died within 30 days of hospitalization were derived from study metadata (when available) and the associated study's manuscripts.

For both tasks, we separated studies into a training set and an external validation set. For the BVN task, the training set consisted of 43 studies (profiled outside Inflammatix) and 3413 patients (1087 with bacterial infection, 1244 with viral infection, and 1082 non-infected). The BVN external validation set consisted of six studies (profiled by Inflammatix) and 293 patients (153 with bacterial infection, 106 with viral infection, and 34 non-infected). For the mortality task, the training set consisted of 33 studies (profiled outside Inflammatix) and 3288 patients (175 30-day mortality events) while the mortality external validation set comprised four studies (profiled by Inflammatix) and 348 patients (80 30-day mortality events). A description of the publicly available studies in our training set appears in Supplementary Table 1.

\subsection{Grouped cross-validation}
Previous analyses by our group~\cite{Mayhew2020} suggested that alternative cross-validation strategies were preferable over conventional k-fold cross-validation (CV) for identifying classifiers able to generalize across heterogeneous patient populations. We use 5-fold grouped CV (full studies are allocated to one and only one of five folds) to rank and select hyperparameter configurations from GS/RS methods and as an objective function in BO.

\subsection{Classifier types and performance assessment}
We evaluated three types of classification models: 1) support vector machines with a radial basis function (RBF) kernel, 2) XGBoost~(XGB~\cite{Chen2016}) and 3) multi-layer perceptrons (MLP). MLP models were trained with the Adam optimizer~\cite{Kingma2014} with mini-batch size fixed at 128.

For the BVN task, we ranked and selected models based on multi-class AUC (mAUC)~\cite{hand2001simple}. For the mortality task, we selected models by binary AUC but report both AUC and average precision to account for class imbalance. To determine performance of models in grouped 5-fold CV, we pooled the model's predicted probabilities for each fold and computed the relevant metric from the pooled probabilities. The top-performing hyperparameter configuration was then trained on the full training set and applied to the external validation set. We computed external validation performance for these top models using their predicted probabilities for the validation samples. We computed 95\% bootstrap confidence intervals for differences in classification performance by sampling predicted probabilities with replacement 5000 times (using the same set of bootstrap sample IDs for both sets of predicted probabilities in the comparison), computing the relevant performance metric on each bootstrap sample, computing the difference between performance metrics for each bootstrap sample in a given comparison, and reporting the 2.5th and 97.5th quantiles of the 5000 differences.

\subsection{Hyperparameter optimization details}
For RBF SVM, we conduct a grid search over configurations of the cost, $C$, and bandwidth hyperparameters, $\gamma$. $C$ values ranged from 1e-03 to 2.15 and $\gamma$ values ranged from 1.12e-04 to 10. We generated RS samples for XGBoost and MLP uniformly and independently of one another from pre-specified ranges or from grids (Suppl. Tables 2 and 3).

For BO, the objective function maps from hyperparameter configurations to 5-fold grouped CV performance of the corresponding classifiers. The two main components of BO are: 1) a model that approximates the objective function, and 2) an acquisition function to propose the next configuration to visit. We use a GP regression model with Gaussian noise to approximate the objective function. To initialize construction of the objective function, we uniformly and independently sample configurations (either 5 or 25) from the hyperparameter space. 

We investigate both the expected improvement and upper confidence bound acquisition functions. We use both standard and automatic relevance determination (ARD) forms of the Matern5/2 covariance function in BO's GP model of the objective (further details in Supplement). We also perform BO in the hyperparameters' native scales (\emph{original} space) or in which continuous and discrete hyperparameter dimensions are searched in the continuous range 0 to 1 and transformed back to their native scales prior to their evaluation (\emph{transformed}). 

\section{Results}
We compared BO and GS/RS approaches for hyperparameter optimization of three types of classifiers 
for two clinical tasks. For the BVN task, we sought classifiers that could achieve high performance in predicting whether a patient had a bacterial or viral infection or was showing a non-infectious inflammatory response. For the mortality task, we sought high-performing classifiers of mortality events within 30 days of hospital admission. Though we considered BO at two initialization budgets (5 and 25 configurations), we did not see substantial differences in performance between classifiers with 5 and 25 initial configurations (Suppl. Table 4, Suppl. Figs. 3-6). We focus on BO results with 25 initial configurations and the expected improvement acquisition function for the remainder of this work (results for all runs in Supplement).

\paragraph{General comparison of classifier performance across tasks and HO approaches}
Across both tasks and HO approaches, we note distinct performance characteristics of the selected classifiers of each type. While RBF SVM classifiers performed similarly to the other two classifier types on the BVN task, they were the worst performers on the mortality task. XGB classifiers selected by either RS or BO demonstrated competitive performance in both tasks and were remarkably consistent in their performance regardless of the number of hyperparameter configurations evaluated for HO. MLPs achieved the highest internal and external validation performance for both acute infection detection and mortality prediction (Table~\ref{tab:ei_unit0_perf}), suggesting potential benefits of learning latent features (hidden layers) for these tasks. We also find that, despite the considerable class imbalance in the mortality task, all classifier types selected by AUC still demonstrated average precision considerably higher than the respective baselines for internal ($\frac{175}{3288} \approx 0.053$) and external ($\frac{80}{348} \approx 0.230$) validation.

\begin{table*}[!htbp]
\caption{Grouped 5-fold CV and external validation (Val.) performance of selected classifiers for the BVN and mortality tasks. BO results used the EI acquisition function and 25 initialization points. The ARD column indicates whether automatic relevance determination was enabled (Y/N) in BO's GP model of the objective function. \textbf{Bold} numbers indicate the best performance for a column. BVN column shows performance in mAUC; mortality column shows AUC performance with average precision in parentheses. $^{*}$Grid specified only 4757 configurations.}
  \centering
\resizebox{0.8\textwidth}{!}{
\begin{tabular}{|c|c|c|c|c|c|c|c|c|}
\hline
\textbf{Model}& \textbf{\shortstack{HO \\Type}} & \textbf{\shortstack{No. of \\Evals.}}  & \textbf{ARD} & \textbf{\shortstack{BVN\\ CV}} & \textbf{\shortstack{BVN\\ Val.}} & \textbf{\shortstack{Mortality\\ CV}} & \textbf{\shortstack{Mortality\\ Val.}} \\ \hline
\hline
\multirow{14}{*}{\textbf{RBF}} & GS & 10 & - & 0.808 & 0.862 & 0.758 (0.182) & 0.736 (0.375) \\
 & GS & 50 & - & 0.814 & 0.853 & 0.797 (0.169) & 0.739 (0.372) \\
 & GS & 100 & - & 0.814 & 0.853 & 0.800 (0.192) & 0.782 (0.533) \\
 & GS & 250 & - & 0.814 & 0.853 & 0.801 (0.191) & 0.749 (0.386) \\
 & GS & 500 & - & 0.815 & 0.853 & 0.801 (0.191) & 0.749 (0.386) \\
 & GS & 1000 & - & 0.815 & 0.853 & 0.839 (0.225) & 0.708 (0.444) \\
 & GS & 5000* & - & 0.815 & 0.853 & 0.839 (0.225) & 0.708 (0.444) \\\cline{2-8}
 & BO & 10 & Y & 0.811 & 0.788 & 0.800 (0.190) & 0.747 (0.383) \\ 
 & BO & 10 & N & 0.815 & 0.851 & 0.800 (0.187) & 0.746 (0.381) \\
 & BO & 50 & Y & 0.816 & 0.852 & 0.801 (0.196) & 0.752 (0.389) \\ 
 & BO & 50 & N & 0.816 & 0.852 & 0.801 (0.194) & 0.749 (0.385) \\ 
 & BO & 100 & Y & 0.816 & 0.852 & 0.800 (0.197) & 0.753 (0.392) \\ 
 & BO & 100 & N & 0.816 & 0.852 & 0.801 (0.196) & 0.752 (0.389) \\
 \hline
\multirow{12}{*}{\textbf{XGB}} & RS & 50 & - & 0.809 & 0.830 & 0.880 (0.315) & 0.819 (0.542) \\
 & RS & 100 & - & 0.813 & 0.827 & 0.885 (0.288) & 0.819 (0.526) \\
 & RS & 250 & - & 0.812 & 0.826 & 0.885 (0.308) & 0.829 (0.556) \\
 & RS & 500 & - & 0.810 & 0.829 & 0.885 (0.320) & 0.826 (0.559) \\
 & RS & 1000 & - & 0.810 & 0.822 & 0.885 (0.311) & 0.822 (0.552) \\
 & RS & 5000 & - & 0.813 & 0.830 & 0.888 (0.310) & 0.823 (0.552) \\
 & RS & 25000 & - & 0.815 & 0.860 & 0.889 (0.303) & 0.816 (0.532) \\\cline{2-8}
 & BO & 50 & Y & 0.818 & 0.865 & 0.887 (0.301) & 0.814 (0.540) \\ 
 & BO & 50 & N & 0.812 & 0.828 & 0.881 (0.275) & 0.817 (0.516) \\ 
 & BO & 100 & Y & 0.811 & 0.825 & 0.885 (0.314) & 0.825 (0.559) \\ 
 & BO & 100 & N & 0.809 & 0.826 & 0.878 (0.288) & 0.817 (0.521) \\ 
 & BO & 250 & Y & 0.818 & 0.865 & 0.886 (0.290) & 0.826 (0.539) \\ 
 & BO & 250 & N & 0.816 & 0.834 & 0.882 (0.272) & 0.802 (0.483) \\
 & BO & 500 & Y & 0.818 & 0.865 & 0.889 (0.346) & 0.827 (0.591) \\ 
 & BO & 500 & N & 0.812 & 0.831 & 0.880 (0.313) & 0.815 (0.538) \\
 \hline
 \multirow{14}{*}{\textbf{MLP}} & RS & 50 & - & 0.818 & 0.860 & 0.763 (0.121) & 0.631 (0.288) \\
 & RS & 100 & - & 0.814 & 0.863 & 0.785 (0.156) & 0.640 (0.301) \\
 & RS & 250 & - & 0.824 & 0.861 & 0.807 (0.211) & 0.625 (0.366) \\
 & RS & 500 & - & 0.819 & 0.859 & 0.853 (0.240) & 0.691 (0.401) \\
 & RS & 1000 & - & 0.835 & \textbf{0.872} & 0.809 (0.158) & 0.637 (0.333) \\
 & RS & 5000 & - & 0.837 & 0.835 & 0.826 (0.249) & 0.796 (0.546) \\
 & RS & 25000 & - & \textbf{0.840} & 0.856 & 0.859 (0.267) & 0.743 (0.428) \\\cline{2-8}
 & BO & 50 & Y & 0.816 & 0.820 & 0.888 (0.340) & 0.823 (0.554) \\ 
 & BO & 50 & N & 0.814 & 0.824 & 0.888 (0.290) & 0.820 (0.564) \\ 
 & BO & 100 & Y & 0.822 & 0.845 & 0.886 (0.296) & \textbf{0.847} (0.631) \\ 
 & BO & 100 & N & 0.828 & 0.854 & 0.884 (0.292) & 0.825 (0.577) \\ 
 & BO & 250 & Y & 0.817 & 0.848 & 0.890 (0.312) & 0.842 (0.614) \\ 
 & BO & 250 & N & 0.832 & 0.832 & 0.889 (0.335) & 0.812 (0.566) \\
 & BO & 500 & Y & 0.837 & 0.855 & \textbf{0.894} (0.304) & 0.835 (0.593) \\ 
 & BO & 500 & N & 0.826 & 0.822 & 0.890 (0.330) & 0.806 (0.561) \\
 \hline
 \end{tabular}
 }
\label{tab:ei_unit0_perf}
\end{table*}

\begin{figure}[htbp]
\begin{minipage}{\textwidth}
  \centering
  \includegraphics[width=\textwidth]{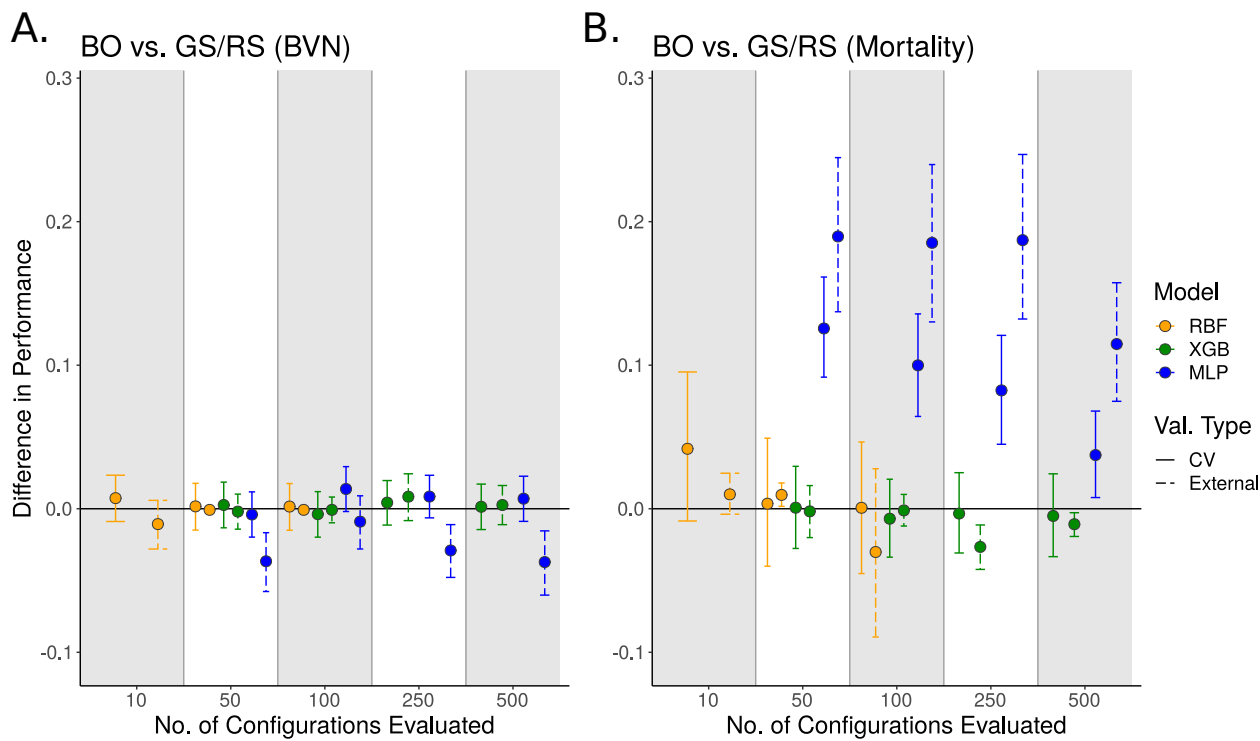}
\end{minipage}
\caption{\textbf{Differences in classification performance of models selected by either BO or GS/RS using BO evaluation budgets.} Performance differences greater than 0 on the BVN (A; mAUC) and mortality (B; AUC) tasks indicate better performance for the BO-selected classifier. Classifiers were selected with the indicated number of hyperparameter configurations evaluated. Automatic relevance determination was not enabled for BO. Points represent observed differences while error bars represent 95\% bootstrap confidence intervals.}
\label{fig:bo_gsrs_direct_comparison}
\end{figure}
 
\paragraph{Evaluation of BO- and GS/RS-selected classifiers at evaluation budgets typical of BO.}
Previous studies have shown that BO can select promising classifiers more efficiently than GS/RS methods. Surprisingly, we find that at smaller numbers of configurations evaluated (more typical of BO), classifiers selected by GS/RS showed similar or better performance in both internal and external validation (Table~\ref{tab:ei_unit0_perf} and Figs.~\ref{fig:bo_gsrs_direct_comparison}) when compared with corresponding BO-selected classifiers. We observed similar trends when using the upper confidence bound acquisition function (Suppl. Figs. 7 and 8, Suppl. Table 5) or the transformed hyperparameter space (Suppl. Figs. 11 and 12, Suppl. Table 6). However, we do note two instances in which BO-selected classifiers exceeded performance of GS/RS-selected classifiers: 1) XGBoost classifiers in external validation for the BVN task and 2) MLP classifiers for the mortality task. While these instances support prior findings of BO's efficiency, our results also suggest that simply committing to a single HO approach could miss models that generalize well and that performance of selected classifiers will depend on the task and classifier type. 

\begin{figure}[htbp]
\begin{minipage}{\textwidth}
  \centering
  \includegraphics[width=\textwidth]{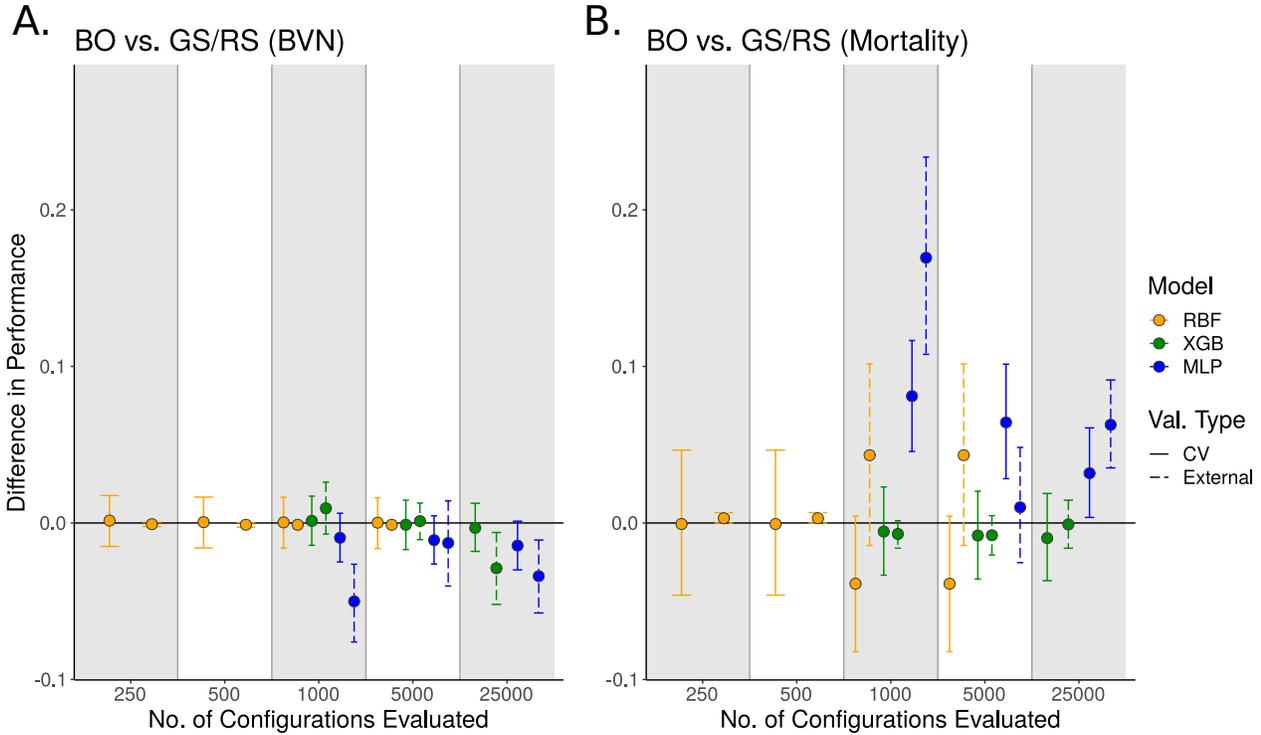}
\end{minipage}
\caption{\textbf{Differences in classification performance of models selected by either BO or GS/RS using GS/RS evaluation budgets.} Run settings and figure layout are the same as in Figure~\ref{fig:bo_gsrs_direct_comparison} except that here, indicated evaluation budgets apply to GS/RS-selected classifiers; BO-selected classifiers are taken from 100-evaluation (RBF) or 500-evaluation (XGB and MLP) runs.}
\label{fig:bo_gsrs_longrange_comparison}
\end{figure}

\paragraph{Evaluation of BO- and GS/RS-selected classifiers at evaluation budgets typical of GS/RS.}
In the previous analysis, we compared BO- and GS/RS-selected classifiers at evaluation budgets typical of BO (i.e. fewer configurations evaluated). In Figure~\ref{fig:bo_gsrs_longrange_comparison}, we compare BO-selected classifiers from their highest evaluation budgets (100 evaluations for RBF and 500 evaluations for XGB and MLP) to classifiers selected by GS/RS at larger evaluation budgets. Interestingly, we find that the BO-selected MLP classifiers for the mortality task continue to outperform their corresponding RS-selected counterparts, even with 25000 configurations evaluated for RS. Similarly, we find that BO-selected XGBoost classifiers exceed external validation performance of RS-selected classifiers on the BVN task up to an evaluation budget of 25000 configurations (though the differences do not persist at 25000 configurations). We observe these differences when conducting BO with the upper confidence bound acquisition function or with a transformed hyperparameter space (Suppl. Figs. 9, 10, 13 and 14). These results indicate the relative efficiency of BO in candidate classifier selection in these two instances but also illustrate the competitiveness of GS/RS-selected classifiers in our setting.

\begin{figure}[htbp]
\begin{minipage}{\textwidth}
  \centering
  \includegraphics[width=\textwidth]{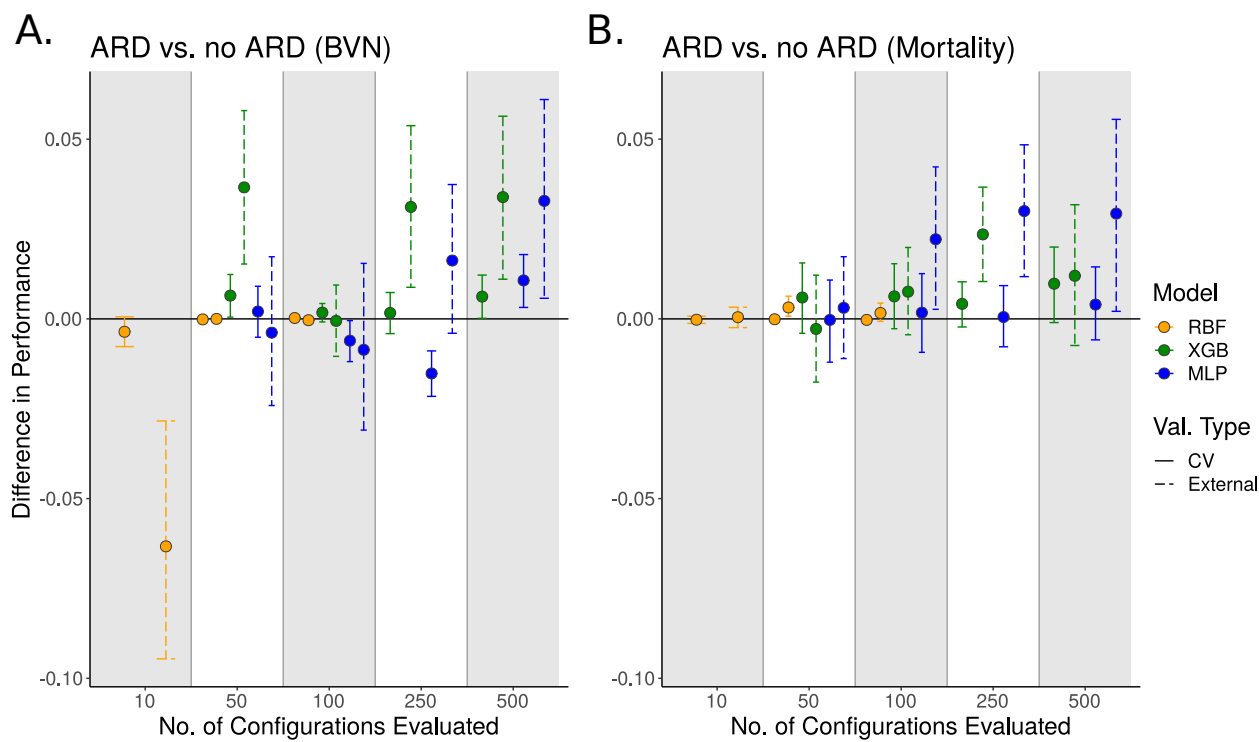}  
\end{minipage}
\caption{\textbf{Differences in classification performance for BO-selected classifiers with or without automatic relevance determination (ARD) enabled.} Performance differences greater than 0 on the BVN (A; mAUC) and mortality (B; AUC) tasks indicate better performance for the classifier selected by ARD-enabled BO. Points represent observed differences while error bars represent 95\% bootstrap confidence intervals.}
\label{fig:bo_ard_compare}
\end{figure}

\paragraph{Assessment of effects on classifier performance of automatic relevance determination in BO.}
For high-dimensional hyperparameter spaces, some hyperparameters may have a greater impact on the model's generalization performance than others. Automatic relevance determination (ARD;~\cite{Neal1996}) in the GP model of BO's objective provides the means to estimate effects of variations in hyperparameter dimensions on the objective's value and has been used in multiple implementations of BO~(e.g. Snoek et al., 2012~\cite{Snoek2012} and BoTorch, \url{https://botorch.org/docs/models}). We directly compare the internal and external validation performance of classifiers selected by BO with and without ARD. In Figure~\ref{fig:bo_ard_compare}, we find that enabling ARD seems to lead to comparable if not slightly better internal validation performance at higher evaluation budgets. Moreover, enabling ARD seems to improve external validation performance for both XGB (BVN task) and MLP classifiers (both tasks). In fact, the highest external validation performance by XGB classifiers on the BVN task is only achieved with ARD enabled (Table~\ref{tab:ei_unit0_perf}). However, these differences in performance are not as evident when using the upper confidence bound acquisition function (Suppl. Fig. 15) or conducting BO in the transformed hyperparameter space (Suppl. Fig. 16). Thus, ARD may not be necessary to select top-performing diagnostic classifiers for these two clinical tasks.

\section{Discussion \& Conclusions}
In this analysis, we compared HO approaches for diagnostic classifier development to determine what approach (if any) led to improvements in: 1) external validation performance or 2) computational efficiency. Consistent with previous findings, we found that BO was able to prioritize candidate classifiers for two tasks relevant to emergency and critical care with a fraction of the configurations evaluated using GS/RS. As embarrassingly parallel approaches like GS/RS can necessitate the use of commodity computing clusters, BO's efficiency makes the approach a potentially cost-effective solution. We also found that external validation performance of BO-selected MLPs for in-hospital mortality was consistently better across a range of HO evaluation budgets than that of GS/RS-selected classifiers, highlighting BO's potential to uncover diagnostic classifiers that generalize better to unseen patients.

However, and in contrast to previous comparisons of HO approaches, our analyses indicated that GS/RS methods could select classifiers for both tasks with evaluation budgets comparable to those used for BO. We also found mixed evidence in support of enabling ARD in the kernel of BO's GP model of the objective function. Thus, while we hoped we would uncover distinct and general differences between HO approaches in order to develop better guidelines about when (or even if) to use one approach over another, we did not identify such clear differences across tasks, classifier types, and optimization settings. Rather, our analysis suggests that both GS/RS and BO approaches should be investigated for classifier development.

We acknowledge limitations of our approach. For our RS runs, we sampled configurations uniformly and independently from pre-defined ranges or grids of values. Other random sampling approaches could've been used in which configurations are generated dependent on the values of previously generated configurations (e.g. Latin hypercube or low-discrepancy Sobol sequences) in order to encourage diversity of the resulting sample~\cite{bergstra2012random}. We felt that the similar performance we observed between BO and GS/RS-selected models using basic variants of GS/RS didn't necessarily justify further analysis with more sophisticated GS/RS variants. A second limitation is that we used a single set of features derived from a previously identified set of 29 gene expression markers. We chose these features based on previous analyses~\cite{Mayhew2020} and consistent with our goal of developing diagnostic classifiers from these specific markers for clinical deployment. We acknowledge our conclusions may not hold with other feature sets. 

Throughout this work, we wanted our hyperparameter optimization to reflect our clinical deployment scenario: that classifiers would likely be evaluated on structured populations (e.g. from a given geographic region) not seen in training. A recent study by Google highlighted this challenge for deployment in healthcare: their AI system for breast cancer screening showed drops in predictive performance when trained on mammograms from the UK and applied to mammograms from the US~\cite{McKinney2020}. However, our survey of ML studies comparing hyperparameter optimization approaches highlighted important differences from our setting in terms of dataset partitioning and, consequently, in the choice of internal validation-based objective function. For example, we found that ML studies primarily focused on larger ($N >\sim$100k) datasets composed mainly of natural images. These benchmarks were often constructed (e.g. MNIST; \url{http://yann.lecun.com/exdb/mnist/}) to satisfy the assumption that the distribution of training and external validation samples are similar if not the same. Internal validation was then performed on subsets of these 'mixed' datasets, with samples from the same structured group in the full dataset appearing in both the training and validation set. However, as patient data is known to be heterogeneous due to biological differences as well as differences in geography, healthcare delivery, and assay technologies used, that assumption of distributional similarity between training and external validation samples is likely to be violated. Indeed, our recent work found that standard k-fold cross-validation gives optimistically biased estimates of generalization error in our setting~\cite{Mayhew2020}, breaking the group structure in left-out folds by randomly distributing patients from the same study into different cross-validation folds (akin to test set contamination). Consequently, in difference to the ML studies we reviewed, we opted for grouped 5-fold cross-validation as our objective function as well as evaluation of performance in external validation to aid model selection.

In conclusion, we find that both GS/RS and BO remain promising avenues for hyperparameter optimization and represent key components in the development of more effective diagnostics for emergency and critical care.

\bibliographystyle{ws-procs11x85}
\bibliography{ms}

\end{document}


\maketitle

\section*{Gene expression features \& data processing}
The features we used in our analyses were based on the expression values of 29 genes previously found to accurately discriminate three different aspects of acute infection: 1) viral vs. bacterial infection (7 genes)~\cite{sweeney2016robust}, 2) infection vs. non-infectious inflammation (11 genes)~\cite{sweeney2015comprehensive}, and 3) high vs. low risk of 30-day mortality (11 genes)~\cite{Sweeney2018}. The 29 genes were grouped into six modules based on positive or negative correlation with infection status or mortality risk: increased expression in viral infections (IFI27, JUP, LAX1), increased expression in bacterial infections (TNIP1, CTSB, HK3, GPAA1), increased (HIF1A, SEPP1, RGS1, C11orf74, CD163, PER1, DEFA4, CIT)  and decreased (LY86, TST, KCNJ2) expression in patients who died within 28 days of hospitalization, increased (CEACAM1, ZDHHC19, C9orf95, GNA15, BATF, C3AR1) and decreased (KIAA1370, TGFBI, MTCH1, RPGRIP1, HLA-DPB1) expression in patients with sepsis. Building on our previous work~\cite{Mayhew2020}, we computed both the geometric means and arithmetic means of these six groups of genes, producing 12 features. We optimized and trained our classifiers on the combination of these 12 features and the expression values of all 29 genes (41 features in total).

We extracted gene expression measurements for all 29 diagnostic markers of interest and normalized these raw expression values for each study. To account for technical and biological heterogeneity across the expression studies, we then performed co-normalization (see~\cite{Mayhew2020} for more detail). We also collected labels for patient infection and mortality status. Labels for one of three classes of the BVN task were determined by multiple-physician adjudication after chart review and not necessarily by positive pathogen identification. Non-infected determinations did not include healthy controls. Binary indicator labels of whether a patient died within 30 days of hospitalization were used for the mortality task. We derived these labels from study metadata (when available) and the associated study's manuscripts.

\begin{table*}
\caption{Publicly-available training studies description. COPD = Chronic Obstructive Pulmonary Disease, ICU = Intensive Care Unit, CAP = Community-Acquired Pneumonia, SIRS = Systemic Inflammatory Response Syndrome, TB = Tuberculosis, SJIA = Systemic Juvenile Idiopathic Arthritis, HRV = human rhinovirus, ED = Emergency Department, A = adult, P = pediatric, B = bacterial infection, V = viral infection, N = noninfected inflammation. Mort. column - total samples with mortality annotation (number of mortality events in parentheses). }
  \centering
\begin{tabular}{|l|l|l|l|l|l|l|}
\hline
\textbf{Study ID  }& \textbf{A/P} & \textbf{Description }  & \textbf{B} & \textbf{V} & \textbf{N} & \textbf{Mort.} \\ \hline
\hline
E-MEXP-3589  & A & Patients hospitalized with COPD exacerbation & 4 & 5 & 14 & 27 (0) \\ \hline
E-MTAB-1548 & A & Surgical patients with sepsis   & 82 & 0 & 58 & 140 (12) \\ \hline
E-MTAB-3162 & A & Dengue patients; Jakarta, Indonesia & 0 & 21 & 0 & 21 (0) \\ \hline
E-MTAB-5273.74 & A & Sepsis due to faecal peritonitis or pneumonia   & 227 & 0 & 0 & - \\ \hline
E-MTAB-5638 & A & Patients with ventilator-associated pneumonia & 0 & 0 & 17 & - \\ \hline
GlueBuffyHCSS & A & Burn and trauma patients & 46 & 0 & 274 & - \\ \hline
GSE103842\_GSE77087 & P & Children with RSV infection & 0 & 103 & 0 & 103 (0) \\ \hline
GSE111368 & A & Adults hospitalised with influenza & 0 & 33 & 0 & - \\ \hline
GSE13015 (GPL6102) & A & Sepsis, many cases from burkholderia & 45 & 0 & 0 & 45 (13) \\ \hline
GSE13015 (GPL6947) & A & Sepsis, many cases from burkholderia  & 15 & 0 & 0 & 15 (7) \\ \hline
GSE21802  & A & Pandemic H1N1 from ICU  & 0 & 10 & 0 & 10 (2) \\ \hline
GSE22098 & A & Patients with active TB and other IDs &  0 & 0 & 71 & 71 (0) \\ \hline
GSE22098 & P & Patients with active TB and other IDs & 111 & 0 & 70 & 181 (0) \\ \hline
GSE25504 (GPL13667) & P & Neonatal sepsis & 9 & 3 & 0 & 12 (1) \\ \hline
GSE25504 (GPL6947) & P & Neonatal sepsis & 20 & 1 & 0 & 21 (1) \\ \hline
GSE27131 & A &  Severe H1N1 & 0 & 7 & 0 & 7 (2) \\ \hline
GSE28750 & A &  Sepsis or post-surgical SIRS & 10 & 0 & 11 & -  \\ \hline
GSE28991 & A &  Sepsis or post-surgical SIRS & 0 & 11 & 0 & 11 (0) \\ \hline
GSE29385\_GSE61821 & A  & Febrile patients positive for H1N1, H3N2 & 0 & 120 & 0 & - \\ \hline
GSE29385\_GSE61821 & P & Febrile patients positive for H1N1, H3N2 & 0 & 8 & 0 & - \\ \hline
GSE32707 & A & Critically ill patients in Brigham \& Women's ICU & 0 & 0 & 44 & 69 (25) \\ \hline
GSE40012\_GSE54514 & A & Bacterial or influenza; A pneumonia or SIRS & 16 & 8 & 12 & 58 (9) \\ \hline
GSE40165 & P & Children and adolescents with Dengue & 0 & 123 & 0 & 123 (0) \\ \hline
GSE40396 & P & Febrile young children & 8 & 22 & 0 & 30 (0) \\ \hline
GSE40586 & A & Community-acquired bacterial meningitis & 15 & 0 & 0 & 15 (2) \\ \hline
GSE42026\_GSE72810 & P & Children with H1N1/09, RSV or bacterial infection & 23 & 51 & 0 & 131 (0) \\ \hline
GSE42834 & A &  Bacterial pneumonia or sarcoidosis   & 14 & 0 & 68 & - \\ \hline
GSE47655 & A & Acute anaphylaxis & 0 & 0 & 6 & 6 (0) \\ \hline
GSE51808 & A & Dengue patients; Bangkok, Thailand & 0 & 28 & 0 & 28 (0) \\ \hline
GSE57065 & A &  Septic shock   &  28 & 0 & 0 & - \\ \hline
GSE57183 & P & SJIA patients & 0 & 0 & 11 & 11 (0) \\ \hline
GSE60244 & A & Lower respiratory tract infections & 22 & 71 & 0 & 118 (0) \\ \hline
GSE63881 & P & Kawasaki Disease & 0 & 0 & 171 & 171 (0) \\ \hline
GSE64456 & P & Febrile infants (60 days of age and younger) & 89 & 111 & 0 & 279 (0) \\ \hline
GSE65682 & A & Suspected but negative for CAP & 0 & 0 & 33 & 106 (23) \\ \hline
GSE66099 & P & Pediatric ICU (sepsis, septic shock, or SIRS) & 109 & 11 & 30 & 229 (28) \\ \hline
GSE67059 & P & Children with HRV infection & 0 & 80 & 0 & 80 (0) \\ \hline
GSE68310 & A & Outpatients with acute respiratory viral infections    & 0 & 104 & 0 & 104 (0) \\ \hline
GSE69528 & A &  Sepsis, many cases from burkholderia    & 83 & 0 & 0 & - \\ \hline
GSE73461 & P & Children with various IDs & 52 & 94 & 162 & 404 (0) \\ \hline
GSE77791 & A & Severe burn shock & 0 & 0 & 30 & - \\ \hline
GSE82050 & A &  Moderate and severe influenza infection    & 0 & 24 & 0 & 24 (0) \\ \hline
GSE95233 & A & Septic shock (unknown pathogen) & - & - & - & 51 (17) \\ \hline
\end{tabular}
\label{tab:training-study-desc}
\end{table*}

\section*{Bayesian optimization details}
We evaluate Bayesian optimization with two different initialization budgets: 5 and 25 initial configurations. Our BO analyses considered both the expected improvement (EI) and upper confidence bound (UCB) acquisition functions. We used the Matern5/2 kernel in BO's GP model of the objective function:
\begin{equation}
    k(r) = \sigma^{2}(1+ \sqrt{5}r + \frac{5}{3}r^{2})\textsf{exp}(-\sqrt{5}r)
\label{eq:matern52}
\end{equation} where 
\begin{equation}
    r = \sqrt{\sum_{h=1}^{H} \frac{(x_{h} - x'_{h})^{2}}{\sigma_{l}^{2}}}.
\label{eq:mat52_sim_noARD}
\end{equation} 
Here $H$ is the dimensionality of the hyperparameter space, $x$ and $x'$ are two configuration vectors being compared, $\sigma^{2}$ is the variance in the objective function's value and $\sigma_{l}$ is a lengthscale parameter that modulates the effect of changes in the hyperparameter dimensions on changes in grouped 5-fold CV performance. In this formulation, there is a single lengthscale parameter for all hyperparameter dimensions. We also assess the automatic relevance determination (ARD) form of the Matern5/2 kernel in which each hyperparameter dimension has their own lengthscale parameter:
\begin{equation}
    r = \sqrt{\sum_{h=1}^{H} \frac{(x_{h} - x'_{h})^{2}}{\sigma_{l,h}^{2}}}
\label{eq:mat52_sim_ARD}
\end{equation}.

\section*{Implementation details}
We used the RBF SVM implementation in Python3's \textsf{scikit-learn} module (version 0.20.1). For XGBoost analyses, we used the \textsf{scikit-learn} API in the \textsf{xgboost} package (version 0.90). All MLP models were implemented using \textsf{Keras} (version 2.1.6) with a \textsf{TensorFlow} (version 1.8.0) back-end. We carried out all GS/RS analyses using our proprietary distributed HO software. In our parallelized implementation of GS/RS, we first generate all configurations and then distribute the configurations to multiple compute nodes in a commodity cluster for evaluation. We executed each GS/RS run using 100 t3a.large instances (2 vCPUs/instance) from Amazon Web Services (AWS). We implemented BO analyses with the \textsf{GPyOpt} package~\cite{gpyopt2016} and carried out BO on a personal laptop with an Intel Core i7-8550U at 1.80GHz 8-core CPU, running Ubuntu 18.04.

\section*{Hyperparameter space details}
\begin{table*}[!htbp]
\vspace{-0.5cm}
\caption{XGBoost hyperparameter space. Shown are the values (for discrete and categorical hyperparameters) and ranges (for continuous hyperparameters) over which optimization runs were conducted. We refer interested readers to \url{https://xgboost.readthedocs.io/en/latest/parameter.html} for a more in-depth description of the different hyperparameters. Where relevant, we indicate when continuous hyperparameters were optimized in \textsf{log} scale and subsequently transformed. All other model parameters were left at default values.}
  \centering
\begin{tabular}{|r|c|c|}
\hline
\textbf{\shortstack{Hyperparameter}} & \textbf{\shortstack{Type}} & \textbf{\shortstack{Values}} \\ \hline
\hline
 booster & categorical & ['gbtree','gblinear','dart'] \\
 \hline
 gamma & continuous & (0,5) \\
 \hline
 learning rate (eta) & continuous & (0.001,0.1) \\
 \hline
 L1 regularization term (alpha) & continuous & (0,1) \\
 \hline
 L2 regularization term (lambda) & continuous & (0,4.5) \\
 \hline
 maximum delta step & discrete & [0,1] \\
 \hline
 maximum tree depth & discrete & [3,4,5,6,7,8,9,10] \\
 \hline
 minimum child weight & discrete & [1,2,3,4,5] \\
 \hline
 number of boosting rounds & continuous (\textsf{log}) & (4.60517,6.907755)  \\
 \hline
 \shortstack{subsample ratio of \\columns for each level} & continuous & (0.7,1) \\
 \hline
 \shortstack{subsample ratio of \\columns for each node} & continuous & (0.5,1) \\
 \hline
 \shortstack{subsample ratio of \\columns for each tree} & continuous & (0.7,1) \\
 \hline
 \shortstack{subsample ratio of \\training instances} & continuous & (0.3,1) \\
 \hline
\end{tabular}
\label{tab:xgb_hyperparams}
\vspace{-1cm}
\end{table*}

\begin{table*}[!htbp]
\caption{MLP hyperparameter space. Shown are the values (for discrete and categorical hyperparameters) and ranges (for continuous hyperparameters) over which optimization runs were conducted. Where relevant, we indicate when continuous hyperparameters were optimized in \textsf{log} scale and subsequently transformed or optimized on a continuous range and converted to the nearest integer (\textsf{int}).}
  \centering
\begin{tabular}{|r|c|c|}
\hline
\textbf{\shortstack{Hyperparameter}} & \textbf{\shortstack{Type}} & \textbf{\shortstack{Values}} \\ \hline
\hline
training iterations & discrete & [50,100,250,500,1000,2500] \\
\hline
number of hidden layers & discrete & [1,2,3,4,5] \\
\hline
\shortstack{number of nodes\\ per hidden layer} & discrete & \shortstack{[2,3,4,5,6,7,8,9,10,11,12,13,\\14,15,16,17,18,19,20,21]} \\
\hline
 activation type & categorical & \shortstack{['ELU','ReLU','sigmoid',\\'tanh','Leaky ReLU']} \\
 \hline
 perform dropout? & categorical & [True,False] \\
 \hline
 \shortstack{dropout rate \\(input layer)} & continuous & (0.0,0.5) \\
 \hline
 \shortstack{dropout rate \\(fully-connected layers)} & continuous & (0.0,0.5) \\
 \hline
 weight regularization type & categorical & ['None','L1','L2'] \\
 \hline
 L1 regularization term & continuous (\textsf{log}) & (-11.512925464970229,0.0) \\
 \hline
 L2 regularization term & continuous (\textsf{log}) & (-11.512925464970229, 0.0) \\
 \hline
 learning rate & continuous (\textsf{log}) & \shortstack{(-11.512925464970229,\\-2.3025850929940455)} \\
 \hline
 weight initialization seed & discrete & [10,11,12,...,10000] \\
 \hline
\end{tabular}
\label{tab:mlp_hyperparams}
\end{table*}

\pagebreak

\begin{figure}[htbp]
\begin{minipage}{\textwidth}
  \centering
  \includegraphics[width=\textwidth]{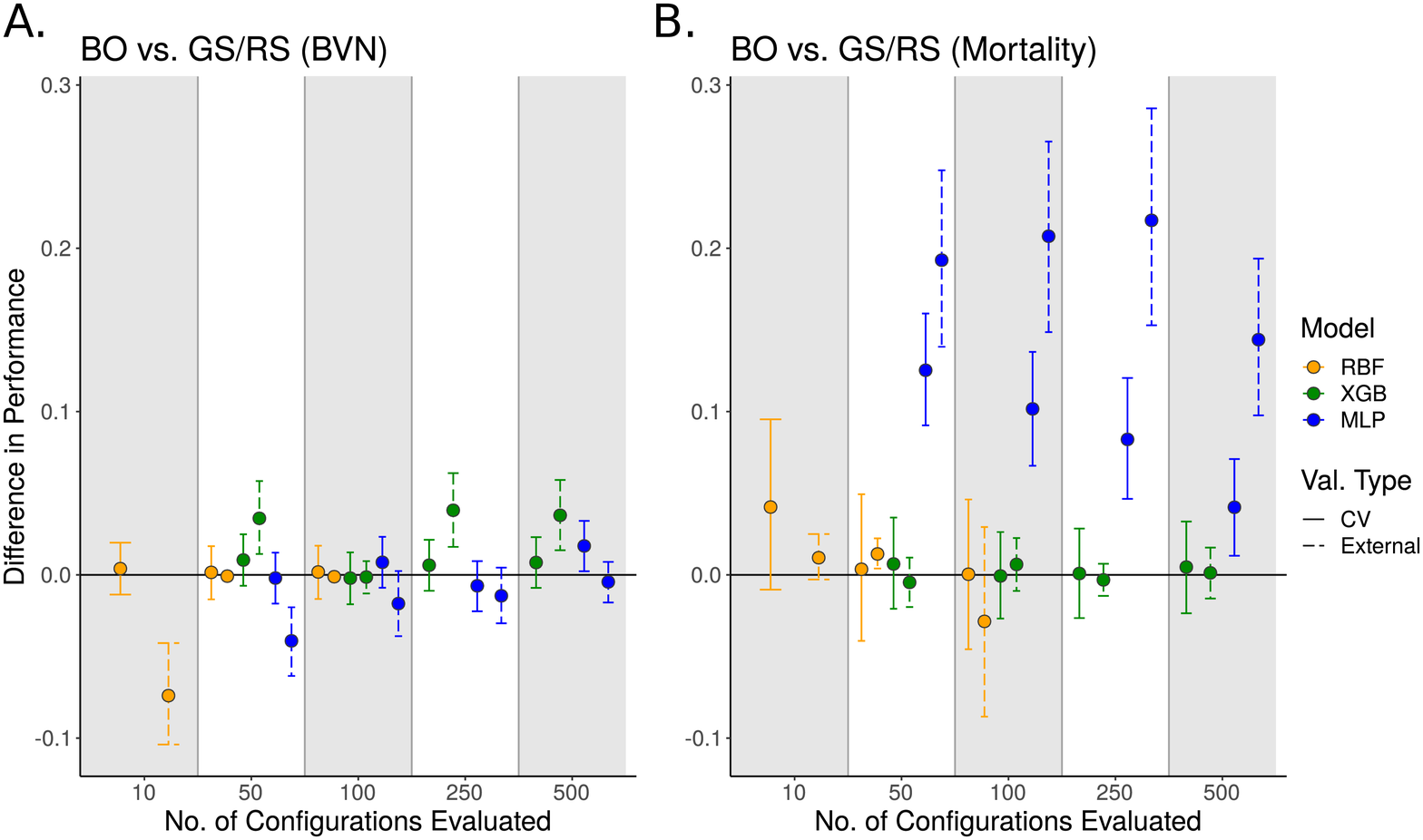}
\end{minipage}
\caption{Differences in classifier performance in internal (CV) and external validation on the BVN (A; mAUC) and mortality (B; AUC) tasks for BO evaluation budgets with the expected improvement acquisition function and 25 initialization configurations in the \emph{original} hyperparameter space. Automatic relevance determination was \emph{enabled} for BO runs. Differences greater than 0 indicate better performance for the BO-selected classifier. Classifiers were selected using either BO or GS/RS with the indicated number of evaluations. Points represent observed differences while error bars represent 95\% bootstrap confidence intervals.}
\label{fig:bo_ei_unit0_ard1_init25_direct_compare}
\end{figure}

\begin{figure}[htbp]
\begin{minipage}{\textwidth}
  \centering
  \includegraphics[width=\textwidth]{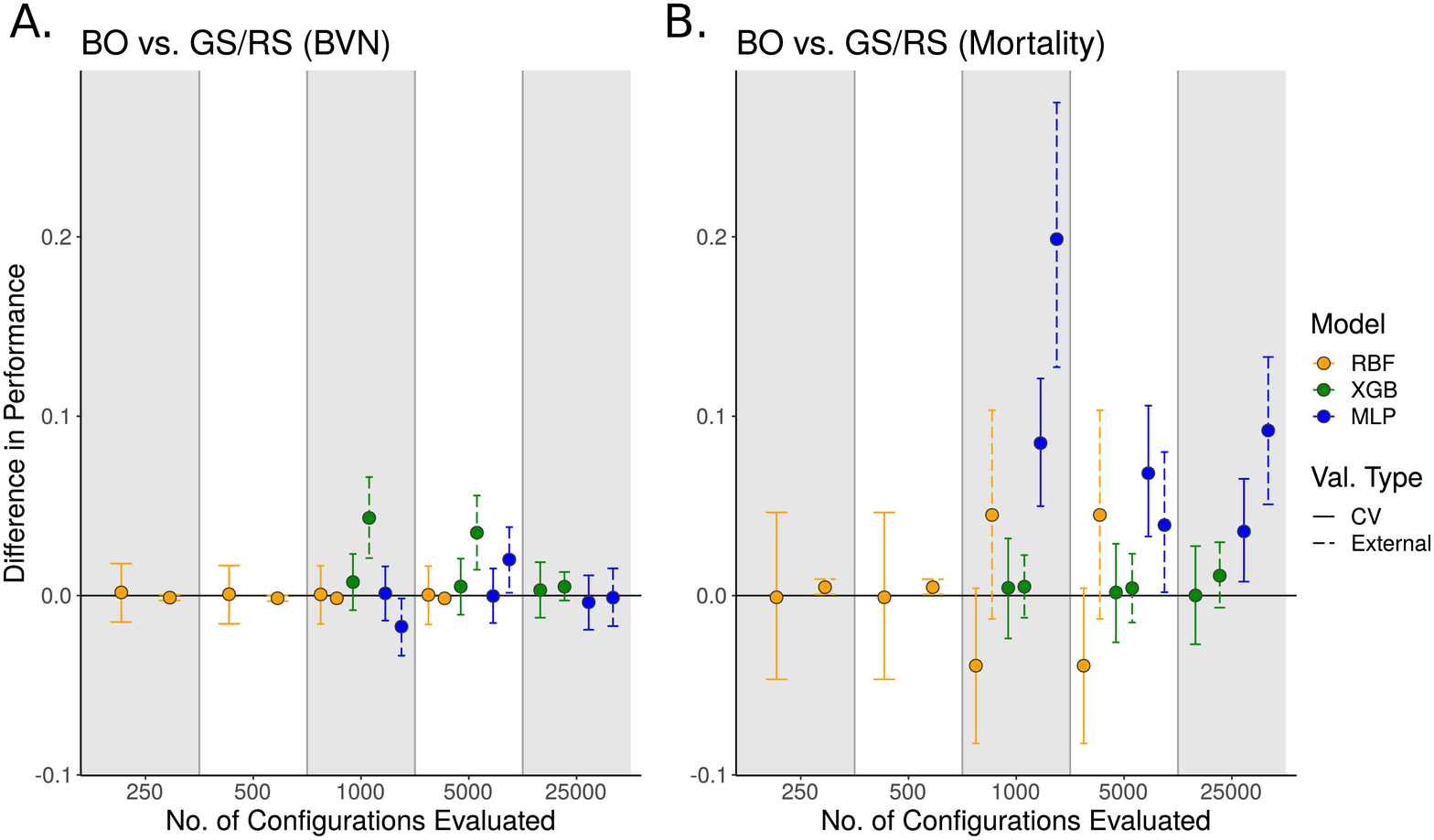}
\end{minipage}
\caption{Differences in classifier performance in internal (CV) and external validation on the BVN (A; mAUC) and mortality (B; AUC) tasks for GS/RS evaluation budgets with the expected improvement acquisition function and 25 initialization configurations in the \emph{original} hyperparameter space. Automatic relevance determination was \emph{enabled} for BO runs. Same run settings and figure layout as in Figure~\ref{fig:bo_ei_unit0_ard1_init25_direct_compare} except that the indicated evaluation budgets only apply to GS/RS-selected classifiers. BO-selected classifiers are taken from 100-evaluation (RBF) or 500-evaluation (XGB and MLP) runs.}
\label{fig:bo_ei_unit0_ard1_init25_long_range_compare}
\end{figure}

\begin{figure}[htbp]
\begin{minipage}{\textwidth}
  \centering
  \includegraphics[width=\textwidth]{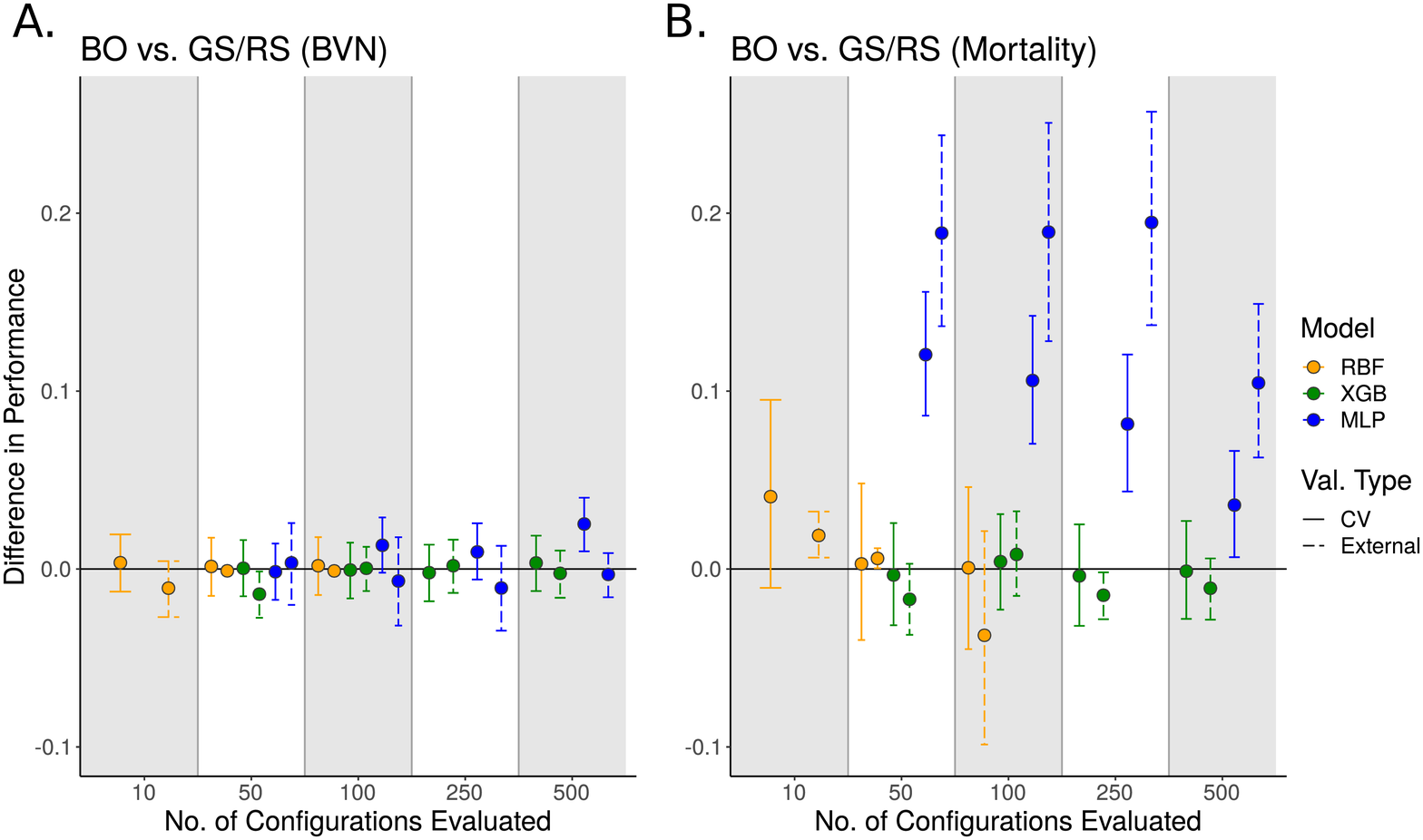}
\end{minipage}
\caption{Differences in classifier performance in internal (CV) and external validation on the BVN (A; mAUC) and mortality (B; AUC) tasks for BO evaluation budgets with the expected improvement acquisition function and 5 initialization configurations in the \emph{original} hyperparameter space. Automatic relevance determination was \emph{disabled} for BO runs. Differences greater than 0 indicate better performance for the BO-selected classifier. Classifiers were selected using either BO or GS/RS with the indicated number of evaluations. Points represent observed differences while error bars represent 95\% bootstrap confidence intervals.}
\label{fig:bo_ei_unit0_ard0_init5_direct_compare}
\end{figure}

\begin{figure}[htbp]
\begin{minipage}{\textwidth}
  \centering
  \includegraphics[width=\textwidth]{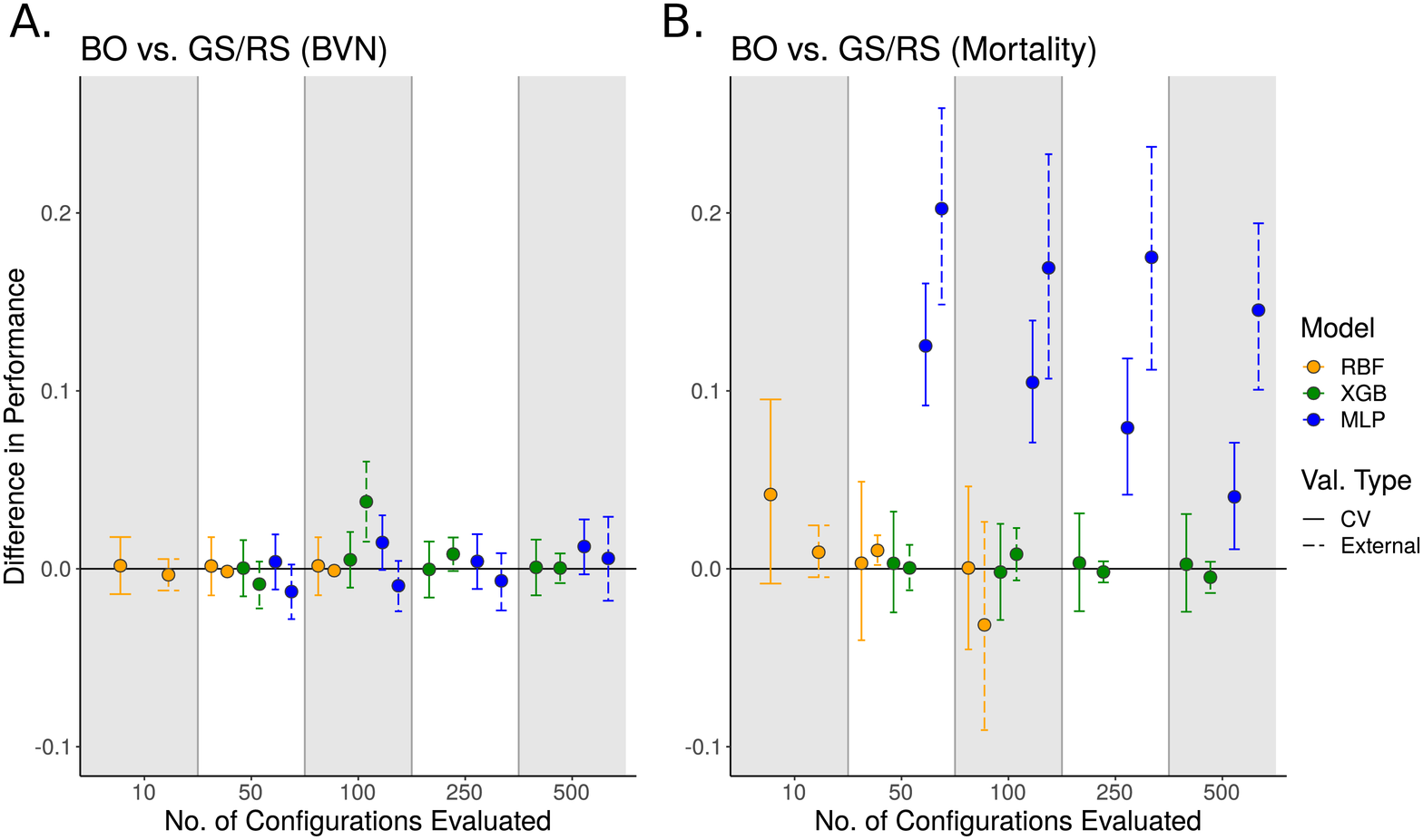}
\end{minipage}
\caption{Differences in classifier performance in internal (CV) and external validation on the BVN (A; mAUC) and mortality (B; AUC) tasks for BO evaluation budgets with the expected improvement acquisition function and 5 initialization configurations in the \emph{original} hyperparameter space. Automatic relevance determination was \emph{enabled} for BO runs. Differences greater than 0 indicate better performance for the BO-selected classifier. Classifiers were selected using either BO or GS/RS with the indicated number of evaluations. Points represent observed differences while error bars represent 95\% bootstrap confidence intervals.}
\label{fig:bo_ei_unit0_ard1_init5_direct_compare}
\end{figure}

\begin{figure}[htbp]
\begin{minipage}{\textwidth}
  \centering
  \includegraphics[width=\textwidth]{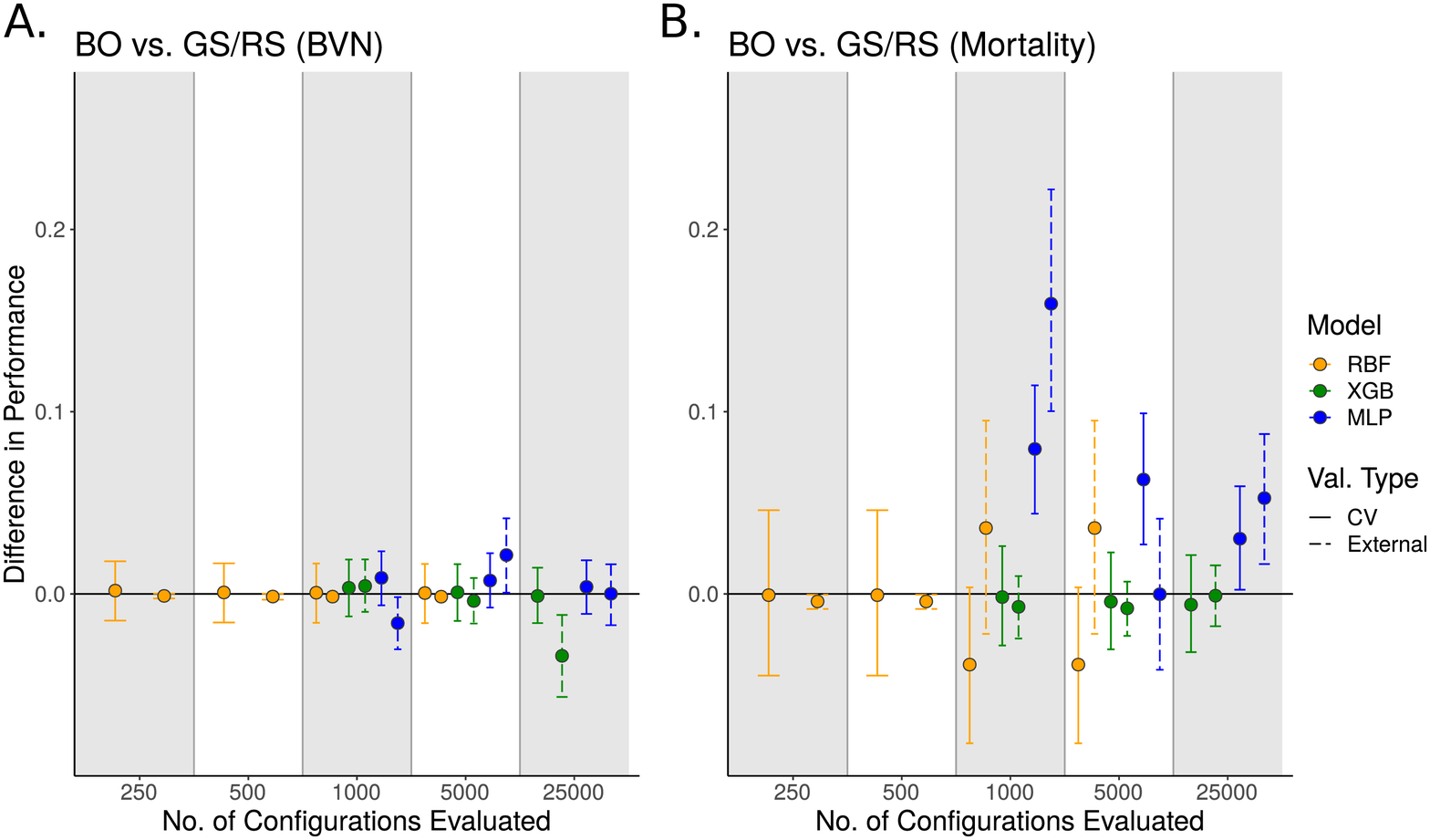}
\end{minipage}
\caption{Differences in classifier performance in internal (CV) and external validation on the BVN (A; mAUC) and mortality (B; AUC) tasks for GS/RS evaluation budgets with the expected improvement acquisition function and 5 initialization configurations in the \emph{original} hyperparameter space. Automatic relevance determination was \emph{disabled} for BO runs. Same run settings and figure layout as in Figure~\ref{fig:bo_ei_unit0_ard0_init5_direct_compare} except that the indicated evaluation budgets only apply to GS/RS-selected classifiers. BO-selected classifiers are taken from 100-evaluation (RBF) or 500-evaluation (XGB and MLP) runs.}
\label{fig:bo_ei_unit0_ard0_init5_long_range_compare}
\end{figure}

\begin{figure}[htbp]
\begin{minipage}{\textwidth}
  \centering
  \includegraphics[width=\textwidth]{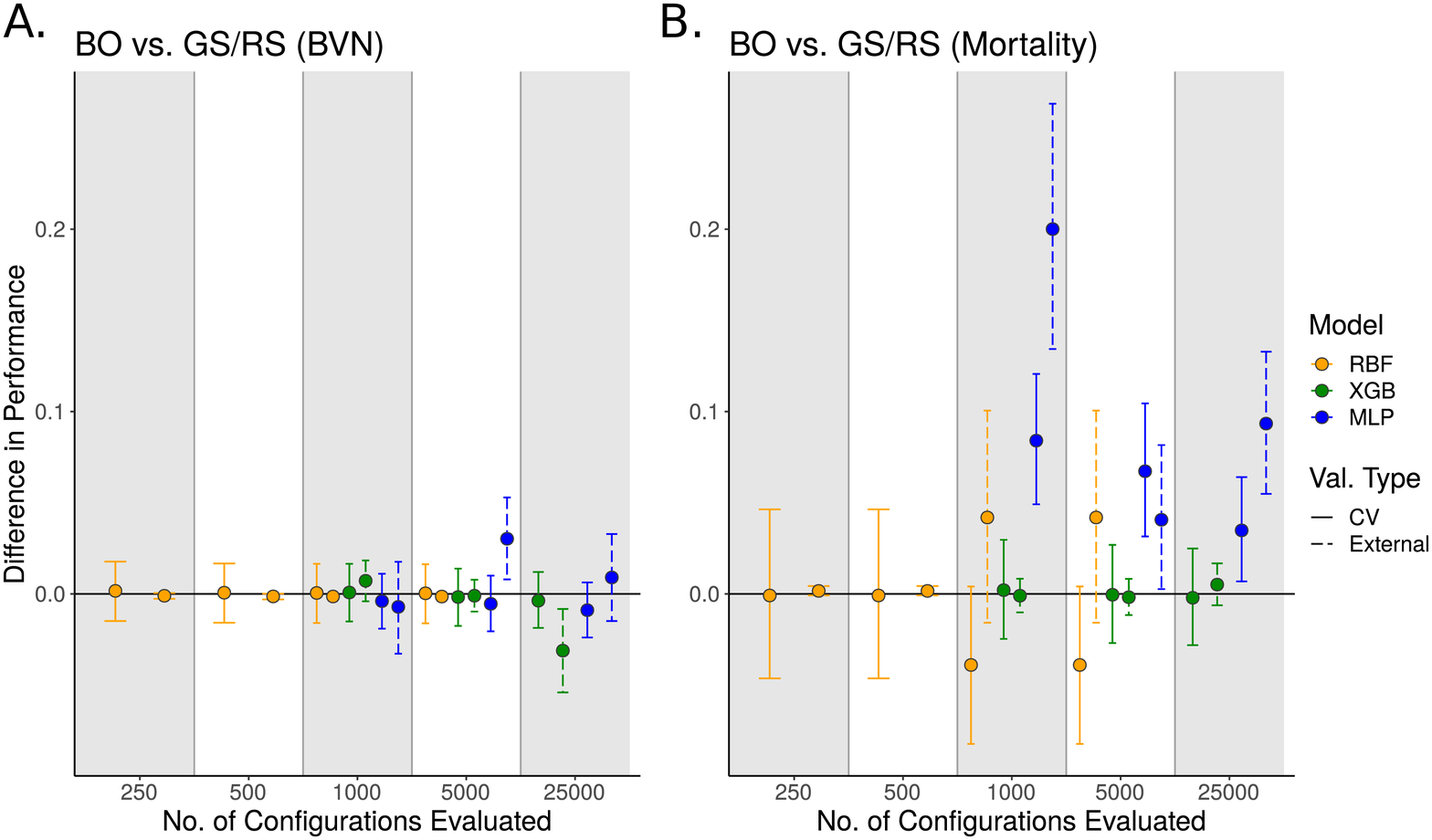}
\end{minipage}
\caption{Differences in classifier performance in internal (CV) and external validation on the BVN (A; mAUC) and mortality (B; AUC) tasks for GS/RS evaluation budgets with the expected improvement acquisition function and 5 initialization configurations in the \emph{original} hyperparameter space. Automatic relevance determination was \emph{enabled} for BO runs. Same run settings and figure layout as in Figure~\ref{fig:bo_ei_unit0_ard1_init5_direct_compare} except that the indicated evaluation budgets only apply to GS/RS-selected classifiers. BO-selected classifiers are taken from 100-evaluation (RBF) or 500-evaluation (XGB and MLP) runs.}
\label{fig:bo_ei_unit0_ard1_init5_long_range_compare}
\end{figure}

\begin{table*}[htbp]
\caption{Grouped 5-fold CV and external validation (Val.) performance of selected classifiers for the BVN and mortality tasks. BO runs were performed in the \emph{original} hyperparameter space with the expected improvement acquisition function and 5 initialization configurations at the indicated evaluation budgets. BVN column shows performance in mAUC; mortality column shows performance in AUC with average precision in parentheses. Note that mortality classifiers were only selected by AUC. \textbf{Bold} numbers indicate the best performance for a column.}
  \centering
\resizebox{0.6\textwidth}{!}{
\begin{tabular}{|c|c|c|c|c|c|c|c|c|}
\hline
\textbf{Model}& \textbf{\shortstack{HO \\Type}} & \textbf{\shortstack{No. of \\Evals.}}  & \textbf{ARD} & \textbf{\shortstack{BVN\\ CV}} & \textbf{\shortstack{BVN\\ Val.}} & \textbf{\shortstack{Mortality\\ CV}} & \textbf{\shortstack{Mortality\\ Val.}} \\ \hline
\hline
\multirow{14}{*}{\textbf{RBF}} & GS & 10 & - & 0.808 & 0.862 & 0.758 (0.182) & 0.736 (0.375) \\
 & GS & 50 & - & 0.814 & 0.853 & 0.797 (0.169) & 0.739 (0.372) \\
 & GS & 100 & - & 0.814 & 0.853 & 0.800 (0.192) & 0.782 (0.533) \\
 & GS & 250 & - & 0.814 & 0.853 & 0.801 (0.191) & 0.749 (0.386) \\
 & GS & 500 & - & 0.815 & 0.853 & 0.801 (0.191) & 0.749 (0.386) \\
 & GS & 1000 & - & 0.815 & 0.853 & 0.839 (0.225) & 0.708 (0.444) \\
 & GS & 5000* & - & 0.815 & 0.853 & 0.839 (0.225) & 0.708 (0.444) \\\cline{2-8}
 & BO & 10 & Y & 0.809 & 0.859 & 0.800 (0.187) & 0.746 (0.381) \\ 
 & BO & 10 & N & 0.811 & 0.851 & 0.799 (0.202) & 0.755 (0.395) \\
 & BO & 50 & Y & 0.816 & 0.851 & 0.800 (0.192) & 0.749 (0.386) \\ 
 & BO & 50 & N & 0.815 & 0.852 & 0.800 (0.186) & 0.745 (0.379) \\ 
 & BO & 100 & Y & 0.816 & 0.852 & 0.800 (0.194) & 0.750 (0.387) \\ 
 & BO & 100 & N & 0.816 & 0.852 & 0.801 (0.188) & 0.745 (0.378) \\
 \hline
 \multirow{12}{*}{\textbf{XGB}} & RS & 50 & - & 0.809 & 0.830 & 0.880 (0.315) & 0.819 (0.542) \\
 & RS & 100 & - & 0.813 & 0.827 & 0.885 (0.288) & 0.819 (0.526) \\
 & RS & 250 & - & 0.812 & 0.826 & 0.885 (0.308) & 0.829 (0.556) \\
 & RS & 500 & - & 0.810 & 0.829 & 0.885 (0.320) & 0.826 (0.559) \\
 & RS & 1000 & - & 0.810 & 0.822 & 0.885 (0.311) & 0.822 (0.552) \\
 & RS & 5000 & - & 0.813 & 0.830 & 0.888 (0.310) & 0.823 (0.552) \\
 & RS & 25000 & - & 0.815 & 0.860 & 0.889 (0.303) & 0.816 (0.532) \\\cline{2-8}
 & BO & 50 & Y & 0.809 & 0.822 & 0.883 (0.301) & 0.819 (0.533) \\ 
 & BO & 50 & N & 0.809 & 0.816 & 0.877 (0.265) & 0.801 (0.499) \\ 
 & BO & 100 & Y & 0.818 & 0.864 & 0.883 (0.318) & 0.827 (0.559) \\ 
 & BO & 100 & N & 0.812 & 0.827 & 0.889 (0.346) & 0.827 (0.591) \\ 
 & BO & 250 & Y & 0.812 & 0.834 & 0.888 (0.317) & 0.827 (0.547) \\ 
 & BO & 250 & N & 0.810 & 0.827 & 0.881 (0.277) & 0.814 (0.525) \\
 & BO & 500 & Y & 0.811 & 0.829 & 0.887 (0.307) & 0.821 (0.552) \\ 
 & BO & 500 & N & 0.814 & 0.826 & 0.883 (0.287) & 0.815 (0.528) \\
 \hline
  \multirow{14}{*}{\textbf{MLP}} & RS & 50 & - & 0.818 & 0.860 & 0.763 (0.121) & 0.631 (0.288) \\
 & RS & 100 & - & 0.814 & 0.863 & 0.785 (0.156) & 0.640 (0.301) \\
 & RS & 250 & - & 0.824 & 0.861 & 0.807 (0.211) & 0.625 (0.366) \\
 & RS & 500 & - & 0.819 & 0.859 & 0.853 (0.240) & 0.691 (0.401) \\
 & RS & 1000 & - & 0.835 & \textbf{0.872} & 0.809 (0.158) & 0.637 (0.333) \\
 & RS & 5000 & - & 0.837 & 0.835 & 0.826 (0.249) & 0.796 (0.546) \\
 & RS & 25000 & - & 0.840 & 0.856 & 0.859 (0.267) & 0.743 (0.428) \\\cline{2-8}
 & BO & 50 & Y & 0.822 & 0.847 & 0.888 (0.336) & 0.833 (0.611) \\ 
 & BO & 50 & N & 0.817 & 0.864 & 0.883 (0.322) & 0.819 (0.582) \\ 
 & BO & 100 & Y & 0.829 & 0.853 & 0.889 (0.322) & 0.809 (0.565) \\ 
 & BO & 100 & N & 0.828 & 0.856 & 0.890 (0.307) & 0.829 (0.616) \\ 
 & BO & 250 & Y & 0.828 & 0.854 & 0.886 (0.324) & 0.800 (0.517) \\ 
 & BO & 250 & N & 0.833 & 0.850 & 0.888 (0.319) & 0.820 (0.576) \\
 & BO & 500 & Y & 0.831 & 0.865 & \textbf{0.893} (0.310) & \textbf{0.837} (0.605) \\ 
 & BO & 500 & N & \textbf{0.844} & 0.856 & 0.889 (0.312) & 0.796 (0.499) \\
 \hline
 \end{tabular}
 }
\label{tab:ei_unit0_init5_perf}
\end{table*}

\begin{table*}[htbp]
\caption{Grouped 5-fold CV and external validation (Val.) performance of selected classifiers for the BVN and mortality tasks. BO runs were performed in the \emph{original} hyperparameter space with the upper confidence bound acquisition function and 25 initialization configurations at the indicated evaluation budgets. BVN column shows performance in mAUC; mortality column shows performance in AUC with average precision in parentheses. Note that mortality classifiers were only selected by AUC. \textbf{Bold} numbers indicate the best performance for a column.}
  \centering
\resizebox{0.6\textwidth}{!}{
\begin{tabular}{|c|c|c|c|c|c|c|c|c|}
\hline
\textbf{Model}& \textbf{\shortstack{HO \\Type}} & \textbf{\shortstack{No. of \\Evals.}}  & \textbf{ARD} & \textbf{\shortstack{BVN\\ CV}} & \textbf{\shortstack{BVN\\ Val.}} & \textbf{\shortstack{Mortality\\ CV}} & \textbf{\shortstack{Mortality\\ Val.}} \\ \hline
\hline
\multirow{14}{*}{\textbf{RBF}} & GS & 10 & - & 0.808 & 0.862 & 0.758 (0.182) & 0.736 (0.375) \\
 & GS & 50 & - & 0.814 & 0.853 & 0.797 (0.169) & 0.739 (0.372) \\
 & GS & 100 & - & 0.814 & 0.853 & 0.800 (0.192) & 0.782 (0.533) \\
 & GS & 250 & - & 0.814 & 0.853 & 0.801 (0.191) & 0.749 (0.386) \\
 & GS & 500 & - & 0.815 & 0.853 & 0.801 (0.191) & 0.749 (0.386) \\
 & GS & 1000 & - & 0.815 & 0.853 & 0.839 (0.225) & 0.708 (0.444) \\
 & GS & 5000* & - & 0.815 & 0.853 & 0.839 (0.225) & 0.708 (0.444) \\\cline{2-8}
 & BO & 10 & Y & 0.814 & 0.851 & 0.800 (0.189) & 0.746 (0.382) \\ 
 & BO & 10 & N & 0.816 & 0.851 & 0.800 (0.196) & 0.752 (0.390) \\
 & BO & 50 & Y & 0.816 & 0.852 & 0.801 (0.195) & 0.752 (0.389) \\ 
 & BO & 50 & N & 0.816 & 0.852 & 0.806 (0.181) & 0.726 (0.500) \\ 
 & BO & 100 & Y & 0.816 & 0.852 & 0.801 (0.193) & 0.748 (0.383) \\ 
 & BO & 100 & N & 0.815 & 0.851 & 0.818 (0.241) & 0.796 (0.555) \\
 \hline
\multirow{12}{*}{\textbf{XGB}} & RS & 50 & - & 0.809 & 0.830 & 0.880 (0.315) & 0.819 (0.542) \\
 & RS & 100 & - & 0.813 & 0.827 & 0.885 (0.288) & 0.819 (0.526) \\
 & RS & 250 & - & 0.812 & 0.826 & 0.885 (0.308) & 0.829 (0.556) \\
 & RS & 500 & - & 0.810 & 0.829 & 0.885 (0.320) & 0.826 (0.559) \\
 & RS & 1000 & - & 0.810 & 0.822 & 0.885 (0.311) & 0.822 (0.552) \\
 & RS & 5000 & - & 0.813 & 0.830 & 0.888 (0.310) & 0.823 (0.552) \\
 & RS & 25000 & - & 0.815 & 0.860 & 0.889 (0.303) & 0.816 (0.532) \\\cline{2-8}
 & BO & 50 & Y & 0.818 & 0.865 & 0.881 (0.318) & 0.816 (0.551) \\ 
 & BO & 50 & N & 0.810 & 0.825 & 0.881 (0.299) & 0.819 (0.550) \\ 
 & BO & 100 & Y & 0.811 & 0.825 & 0.888 (0.326) & 0.821 (0.587) \\ 
 & BO & 100 & N & 0.808 & 0.812 & 0.882 (0.316) & 0.822 (0.563) \\ 
 & BO & 250 & Y & 0.818 & 0.865 & 0.890 (0.337) & 0.821 (0.581) \\ 
 & BO & 250 & N & 0.811 & 0.823 & 0.882 (0.310) & 0.821 (0.551) \\
 & BO & 500 & Y & 0.818 & 0.865 & 0.888 (0.319) & 0.821 (0.554) \\ 
 & BO & 500 & N & 0.816 & 0.826 & 0.885 (0.328) & 0.825 (0.550) \\
 \hline
 \multirow{14}{*}{\textbf{MLP}} & RS & 50 & - & 0.818 & 0.860 & 0.763 (0.121) & 0.631 (0.288) \\
 & RS & 100 & - & 0.814 & 0.863 & 0.785 (0.156) & 0.640 (0.301) \\
 & RS & 250 & - & 0.824 & 0.861 & 0.807 (0.211) & 0.625 (0.366) \\
 & RS & 500 & - & 0.819 & 0.859 & 0.853 (0.240) & 0.691 (0.401) \\
 & RS & 1000 & - & 0.835 & \textbf{0.872} & 0.809 (0.158) & 0.637 (0.333) \\
 & RS & 5000 & - & 0.837 & 0.835 & 0.826 (0.249) & 0.796 (0.546) \\
 & RS & 25000 & - & 0.840 & 0.856 & 0.859 (0.267) & 0.743 (0.428) \\\cline{2-8}
 & BO & 50 & Y & 0.820 & 0.860 & 0.886 (0.335) & 0.828 (0.582) \\ 
 & BO & 50 & N & 0.826 & 0.834 & 0.889 (0.306) & 0.833 (0.595) \\ 
 & BO & 100 & Y & 0.825 & 0.850 & 0.884 (0.291) & 0.843 (0.618) \\ 
 & BO & 100 & N & 0.817 & 0.841 & 0.890 (0.340) & 0.817 (0.558) \\ 
 & BO & 250 & Y & 0.831 & 0.846 & 0.891 (0.310) & 0.820 (0.570) \\ 
 & BO & 250 & N & 0.826 & 0.858 & \textbf{0.893} (0.335) & 0.831 (0.600) \\
 & BO & 500 & Y & 0.830 & 0.845 & 0.886 (0.336) & 0.824 (0.570) \\ 
 & BO & 500 & N & \textbf{0.844} & 0.824 & 0.892 (0.343) & \textbf{0.844} (0.636) \\
 \hline
 \end{tabular}
 }
\label{tab:ucb_unit0_init25_perf}
\end{table*}

\begin{table*}[htbp]
\caption{Grouped 5-fold CV and external validation (Val.) performance of selected classifiers for the BVN and mortality tasks. BO runs were performed in the \emph{transformed} hyperparameter space with the expected improvement acquisition function and 25 initialization configurations at the indicated evaluation budgets. BVN column shows performance in mAUC; mortality column shows performance in AUC with average precision in parentheses. Note that mortality classifiers were only selected by AUC. \textbf{Bold} numbers indicate the best performance for a column.}
  \centering
\resizebox{0.6\textwidth}{!}{
\begin{tabular}{|c|c|c|c|c|c|c|c|c|}
\hline
\textbf{Model}& \textbf{\shortstack{HO \\Type}} & \textbf{\shortstack{No. of \\Evals.}}  & \textbf{ARD} & \textbf{\shortstack{BVN\\ CV}} & \textbf{\shortstack{BVN\\ Val.}} & \textbf{\shortstack{Mortality\\ CV}} & \textbf{\shortstack{Mortality\\ Val.}} \\ \hline
\hline
\multirow{14}{*}{\textbf{RBF}} & GS & 10 & - & 0.808 & 0.862 & 0.758 (0.182) & 0.736 (0.375) \\
 & GS & 50 & - & 0.814 & 0.853 & 0.797 (0.169) & 0.739 (0.372) \\
 & GS & 100 & - & 0.814 & 0.853 & 0.800 (0.192) & 0.782 (0.533) \\
 & GS & 250 & - & 0.814 & 0.853 & 0.801 (0.191) & 0.749 (0.386) \\
 & GS & 500 & - & 0.815 & 0.853 & 0.801 (0.191) & 0.749 (0.386) \\
 & GS & 1000 & - & 0.815 & 0.853 & 0.839 (0.225) & 0.708 (0.444) \\
 & GS & 5000* & - & 0.815 & 0.853 & 0.839 (0.225) & 0.708 (0.444) \\\cline{2-8}
 & BO & 10 & Y & 0.813 & 0.851 & 0.817 (0.217) & 0.678 (0.452) \\ 
 & BO & 10 & N & 0.815 & 0.852 & 0.832 (0.209) & 0.772 (0.532) \\
 & BO & 50 & Y & 0.815 & 0.851 & 0.800 (0.189) & 0.748 (0.384) \\ 
 & BO & 50 & N & 0.816 & 0.852 & 0.802 (0.200) & 0.735 (0.509) \\ 
 & BO & 100 & Y & 0.816 & 0.852 & 0.810 (0.236) & 0.727 (0.498) \\ 
 & BO & 100 & N & 0.816 & 0.852 & 0.801 (0.195) & 0.749 (0.385) \\
 \hline
\multirow{12}{*}{\textbf{XGB}} & RS & 50 & - & 0.809 & 0.830 & 0.880 (0.315) & 0.819 (0.542) \\
 & RS & 100 & - & 0.813 & 0.827 & 0.885 (0.288) & 0.819 (0.526) \\
 & RS & 250 & - & 0.812 & 0.826 & 0.885 (0.308) & 0.829 (0.556) \\
 & RS & 500 & - & 0.810 & 0.829 & 0.885 (0.320) & 0.826 (0.559) \\
 & RS & 1000 & - & 0.810 & 0.822 & 0.885 (0.311) & 0.822 (0.552) \\
 & RS & 5000 & - & 0.813 & 0.830 & 0.888 (0.310) & 0.823 (0.552) \\
 & RS & 25000 & - & 0.815 & 0.860 & 0.889 (0.303) & 0.816 (0.532) \\\cline{2-8}
 & BO & 50 & Y & 0.809 & 0.828 & 0.883 (0.315) & 0.824 (0.565) \\ 
 & BO & 50 & N & 0.811 & 0.827 & 0.882 (0.298) & 0.815 (0.517) \\ 
 & BO & 100 & Y & 0.810 & 0.826 & 0.885 (0.312) & 0.824 (0.550) \\ 
 & BO & 100 & N & 0.811 & 0.830 & 0.883 (0.293) & 0.826 (0.534) \\ 
 & BO & 250 & Y & 0.810 & 0.820 & 0.887 (0.318) & 0.816 (0.546) \\ 
 & BO & 250 & N & 0.811 & 0.817 & 0.886 (0.301) & 0.823 (0.538) \\
 & BO & 500 & Y & 0.814 & 0.823 & 0.889 (0.307) & 0.823 (0.557) \\ 
 & BO & 500 & N & 0.808 & 0.826 & 0.883 (0.288) & 0.811 (0.513) \\
 \hline
 \multirow{14}{*}{\textbf{MLP}} & RS & 50 & - & 0.818 & 0.860 & 0.763 (0.121) & 0.631 (0.288) \\
 & RS & 100 & - & 0.814 & 0.863 & 0.785 (0.156) & 0.640 (0.301) \\
 & RS & 250 & - & 0.824 & 0.861 & 0.807 (0.211) & 0.625 (0.366) \\
 & RS & 500 & - & 0.819 & 0.859 & 0.853 (0.240) & 0.691 (0.401) \\
 & RS & 1000 & - & 0.835 & 0.872 & 0.809 (0.158) & 0.637 (0.333) \\
 & RS & 5000 & - & 0.837 & 0.835 & 0.826 (0.249) & 0.796 (0.546) \\
 & RS & 25000 & - & \textbf{0.840} & 0.856 & 0.859 (0.267) & 0.743 (0.428) \\\cline{2-8}
 & BO & 50 & Y & 0.815 & 0.840 & 0.890 (0.341) & 0.821 (0.568) \\ 
 & BO & 50 & N & 0.819 & 0.873 & 0.888 (0.322) & 0.824 (0.588) \\ 
 & BO & 100 & Y & 0.822 & \textbf{0.878} & \textbf{0.893} (0.323) & 0.836 (0.596) \\ 
 & BO & 100 & N & 0.817 & 0.858 & 0.889 (0.325) & \textbf{0.837} (0.602) \\ 
 & BO & 250 & Y & 0.824 & 0.863 & 0.889 (0.306) & 0.831 (0.615) \\ 
 & BO & 250 & N & 0.831 & 0.856 & 0.892 (0.306) & 0.836 (0.608) \\
 & BO & 500 & Y & 0.833 & 0.858 & 0.890 (0.319) & 0.833 (0.599) \\ 
 & BO & 500 & N & 0.832 & 0.843 & 0.892 (0.319) & 0.809 (0.540) \\
 \hline
 \end{tabular}
 }
\label{tab:ei_unit1_init25_perf}
\end{table*}

\begin{figure}[htbp]
\begin{minipage}{\textwidth}
  \centering
  \includegraphics[width=\textwidth]{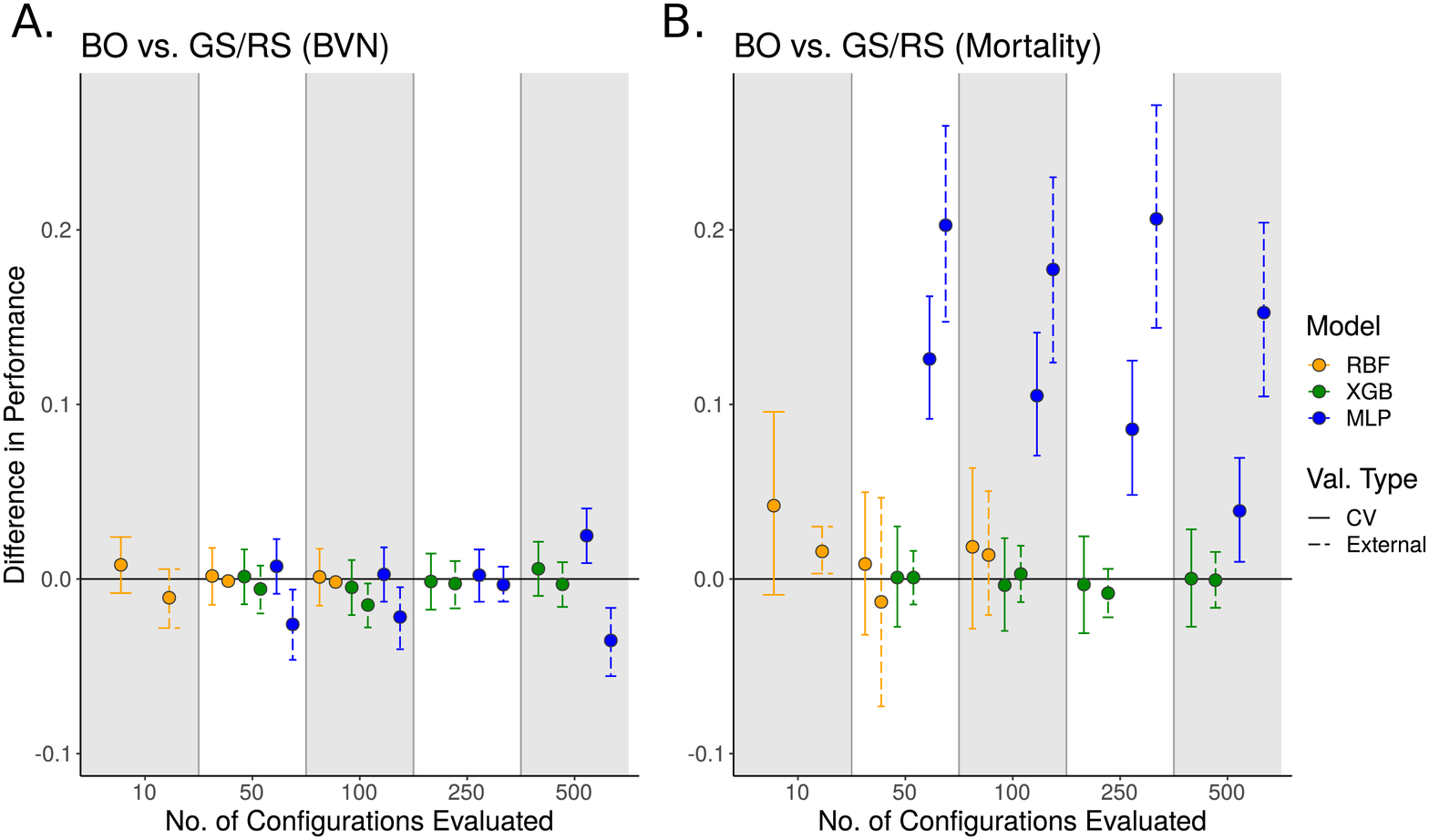}
\end{minipage}
\caption{Differences in classifier performance in internal (CV) and external validation on the BVN (A; mAUC) and mortality (B; AUC) tasks for BO evaluation budgets with the upper confidence bound acquisition function and 25 initialization configurations in the \emph{original} hyperparameter space. Automatic relevance determination was \emph{disabled} for BO runs. Differences greater than 0 indicate better performance for the BO-selected classifier. Classifiers were selected using either BO or GS/RS with the indicated number of evaluations. Points represent observed differences while error bars represent 95\% bootstrap confidence intervals.}
\label{fig:bo_ucb_unit0_ard0_init25_direct_compare}
\end{figure}

\begin{figure}[htbp]
\begin{minipage}{\textwidth}
  \centering
  \includegraphics[width=\textwidth]{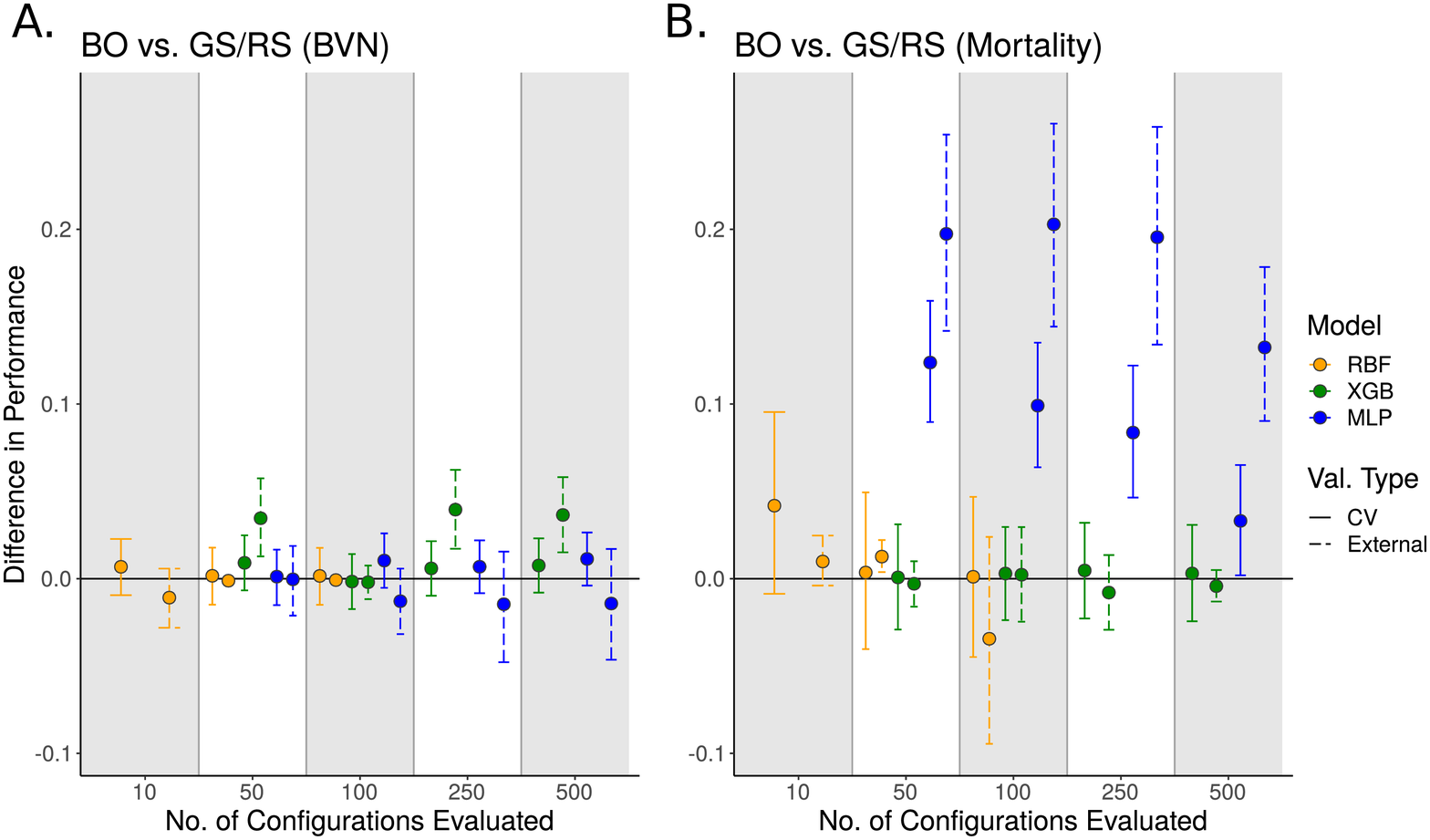}
\end{minipage}
\caption{Differences in classifier performance in internal (CV) and external validation on the BVN (A; mAUC) and mortality (B; AUC) tasks for BO evaluation budgets with the upper confidence bound acquisition function and 25 initialization configurations in the \emph{original} hyperparameter space. Automatic relevance determination was \emph{enabled} for BO runs. Differences greater than 0 indicate better performance for the BO-selected classifier. Classifiers were selected using either BO or GS/RS with the indicated number of evaluations. Points represent observed differences while error bars represent 95\% bootstrap confidence intervals.}
\label{fig:bo_ucb_unit0_ard1_init25_direct_compare}
\end{figure}

\begin{figure}[htbp]
\begin{minipage}{\textwidth}
  \centering
  \includegraphics[width=\textwidth]{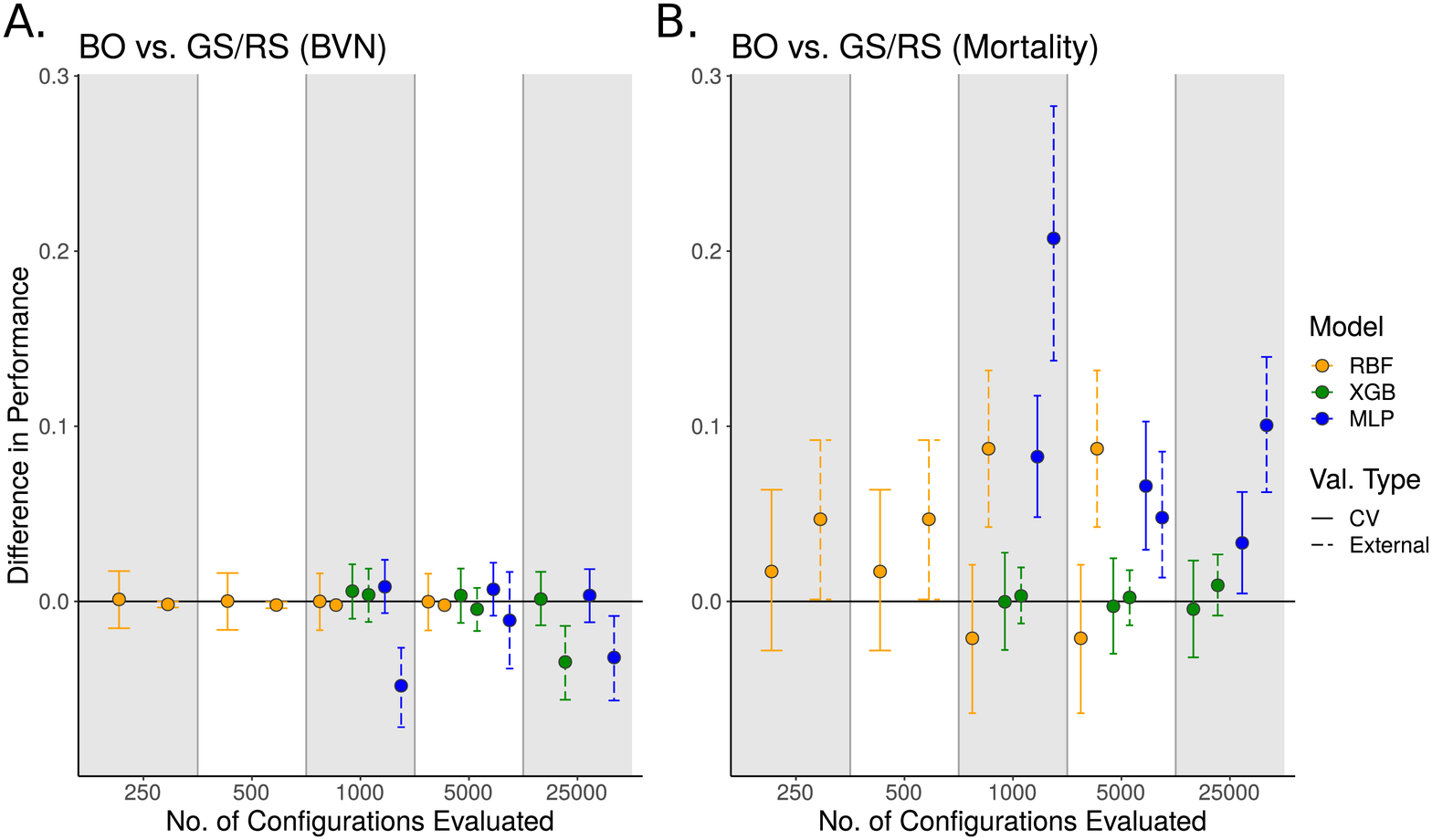}
\end{minipage}
\caption{Differences in classifier performance in internal (CV) and external validation on the BVN (A; mAUC) and mortality (B; AUC) tasks for GS/RS evaluation budgets with the upper confidence bound acquisition function and 25 initialization configurations in the \emph{original} hyperparameter space. Automatic relevance determination was \emph{disabled} for BO runs. Same run settings and figure layout as in Figure~\ref{fig:bo_ucb_unit0_ard0_init25_direct_compare} except that the indicated evaluation budgets only apply to GS/RS-selected classifiers. BO-selected classifiers are taken from 100-evaluation (RBF) or 500-evaluation (XGB and MLP) runs.}
\label{fig:bo_ucb_unit0_ard0_init25_long_range_compare}
\end{figure}

\begin{figure}[htbp]
\begin{minipage}{\textwidth}
  \centering
  \includegraphics[width=\textwidth]{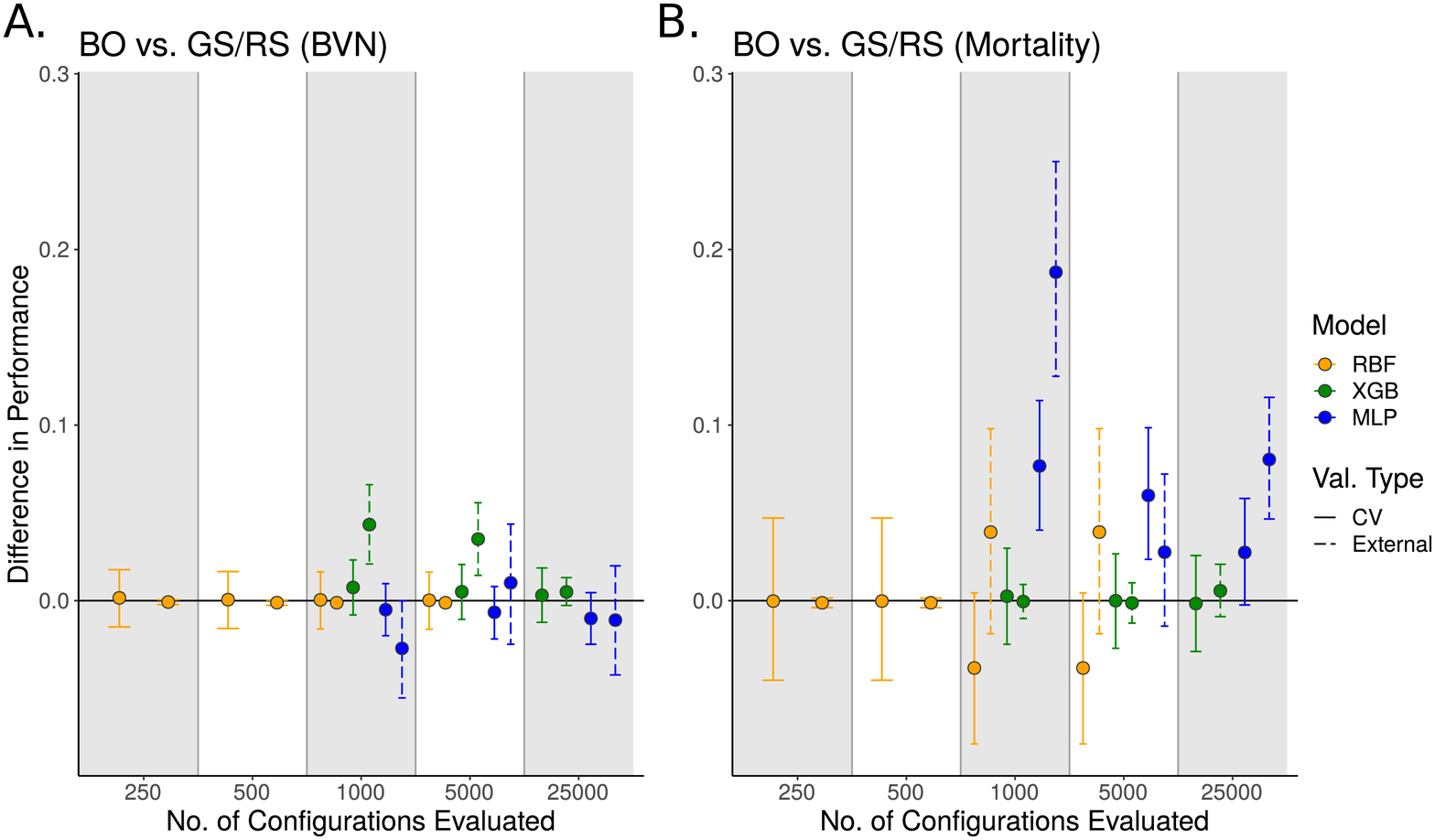}
\end{minipage}
\caption{Differences in classifier performance in internal (CV) and external validation on the BVN (A; mAUC) and mortality (B; AUC) tasks for GS/RS evaluation budgets with the upper confidence bound acquisition function and 25 initialization configurations in the \emph{original} hyperparameter space. Automatic relevance determination was \emph{enabled} for BO runs. Same run settings and figure layout as in Figure~\ref{fig:bo_ucb_unit0_ard1_init25_direct_compare} except that the indicated evaluation budgets only apply to GS/RS-selected classifiers. BO-selected classifiers are taken from 100-evaluation (RBF) or 500-evaluation (XGB and MLP) runs.}
\label{fig:bo_ucb_unit0_ard1_init25_long_range_compare}
\end{figure}

\begin{figure}[htbp]
\begin{minipage}{\textwidth}
  \centering
  \includegraphics[width=\textwidth]{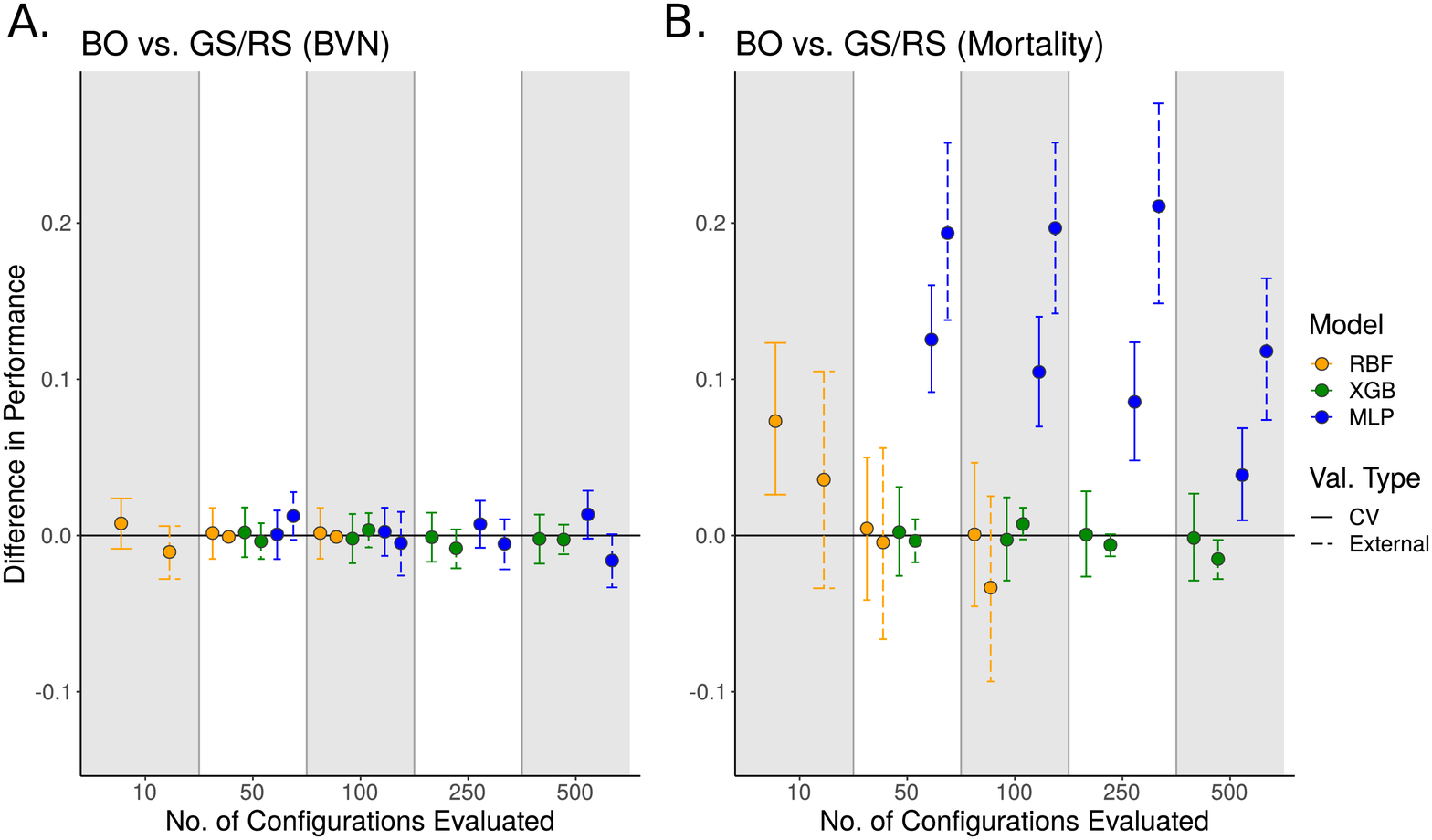}
\end{minipage}
\caption{Differences in classifier performance in internal (CV) and external validation on the BVN (A; mAUC) and mortality (B; AUC) tasks for BO evaluation budgets with the expected improvement acquisition function and 25 initialization configurations in the \emph{transformed} hyperparameter space. Automatic relevance determination was \emph{disabled} for BO runs. Differences greater than 0 indicate better performance for the BO-selected classifier. Classifiers were selected using either BO or GS/RS with the indicated number of evaluations. Points represent observed differences while error bars represent 95\% bootstrap confidence intervals.}
\label{fig:bo_ei_unit1_ard0_init25_direct_compare}
\end{figure}

\begin{figure}[htbp]
\begin{minipage}{\textwidth}
  \centering
  \includegraphics[width=\textwidth]{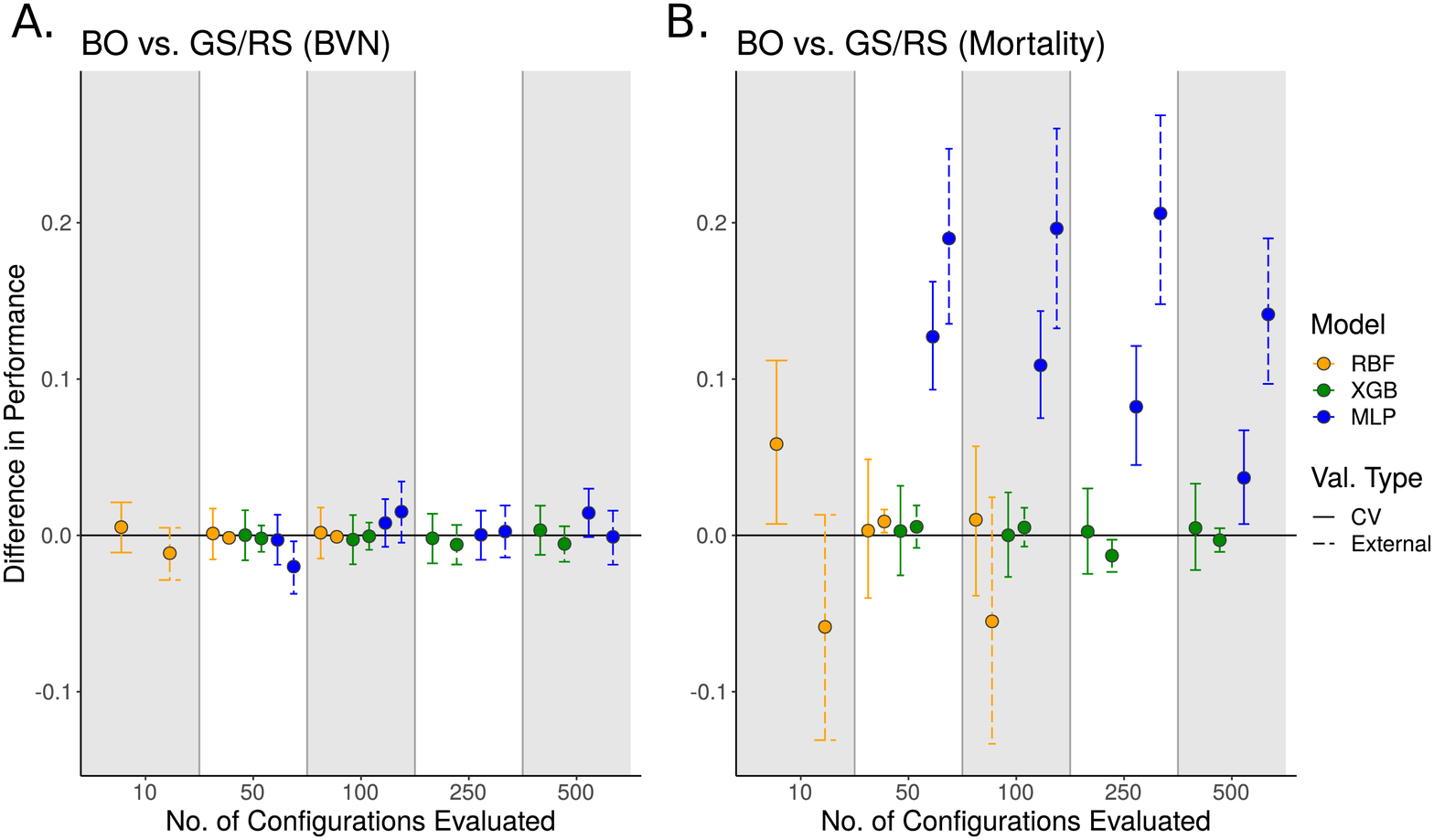}
\end{minipage}
\caption{Differences in classifier performance in internal (CV) and external validation on the BVN (A; mAUC) and mortality (B; AUC) tasks for BO evaluation budgets with the expected improvement acquisition function and 25 initialization configurations in the \emph{transformed} hyperparameter space. Automatic relevance determination was \emph{enabled} for BO runs. Differences greater than 0 indicate better performance for the BO-selected classifier. Classifiers were selected using either BO or GS/RS with the indicated number of evaluations. Points represent observed differences while error bars represent 95\% bootstrap confidence intervals.}
\label{fig:bo_ei_unit1_ard1_init25_direct_compare}
\end{figure}

\begin{figure}[htbp]
\begin{minipage}{\textwidth}
  \centering
  \includegraphics[width=\textwidth]{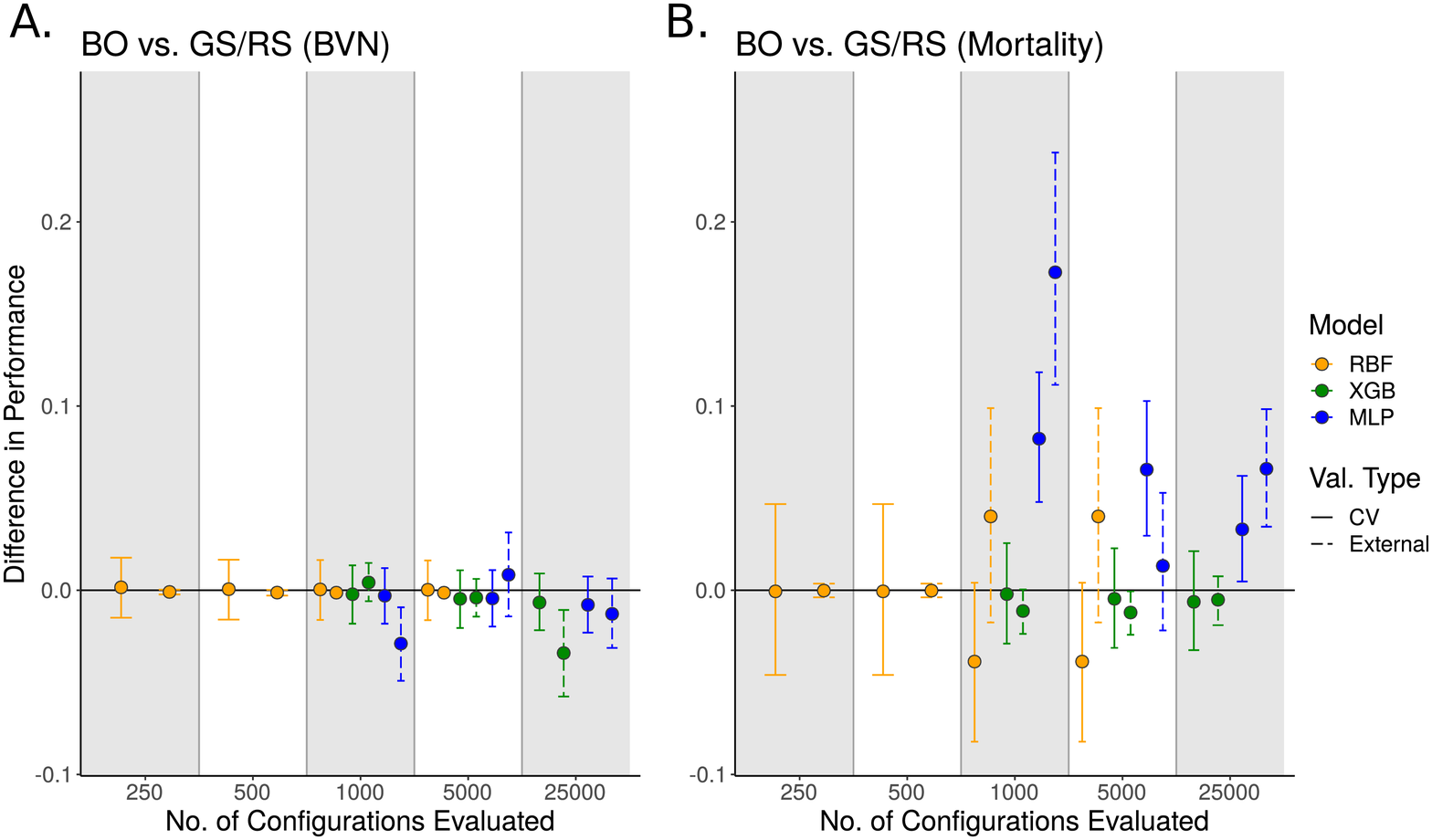}
\end{minipage}
\caption{Differences in classifier performance in internal (CV) and external validation on the BVN (A; mAUC) and mortality (B; AUC) tasks for GS/RS evaluation budgets with the expected improvement acquisition function and 25 initialization configurations in the \emph{transformed} hyperparameter space. Automatic relevance determination was \emph{disabled} for BO runs. Same run settings and figure layout as in Figure~\ref{fig:bo_ei_unit1_ard0_init25_direct_compare} except that the indicated evaluation budgets only apply to GS/RS-selected classifiers. BO-selected classifiers are taken from 100-evaluation (RBF) or 500-evaluation (XGB and MLP) runs.}
\label{fig:bo_ei_unit1_ard0_init25_long_range_compare}
\end{figure}

\begin{figure}[htbp]
\begin{minipage}{\textwidth}
  \centering
  \includegraphics[width=\textwidth]{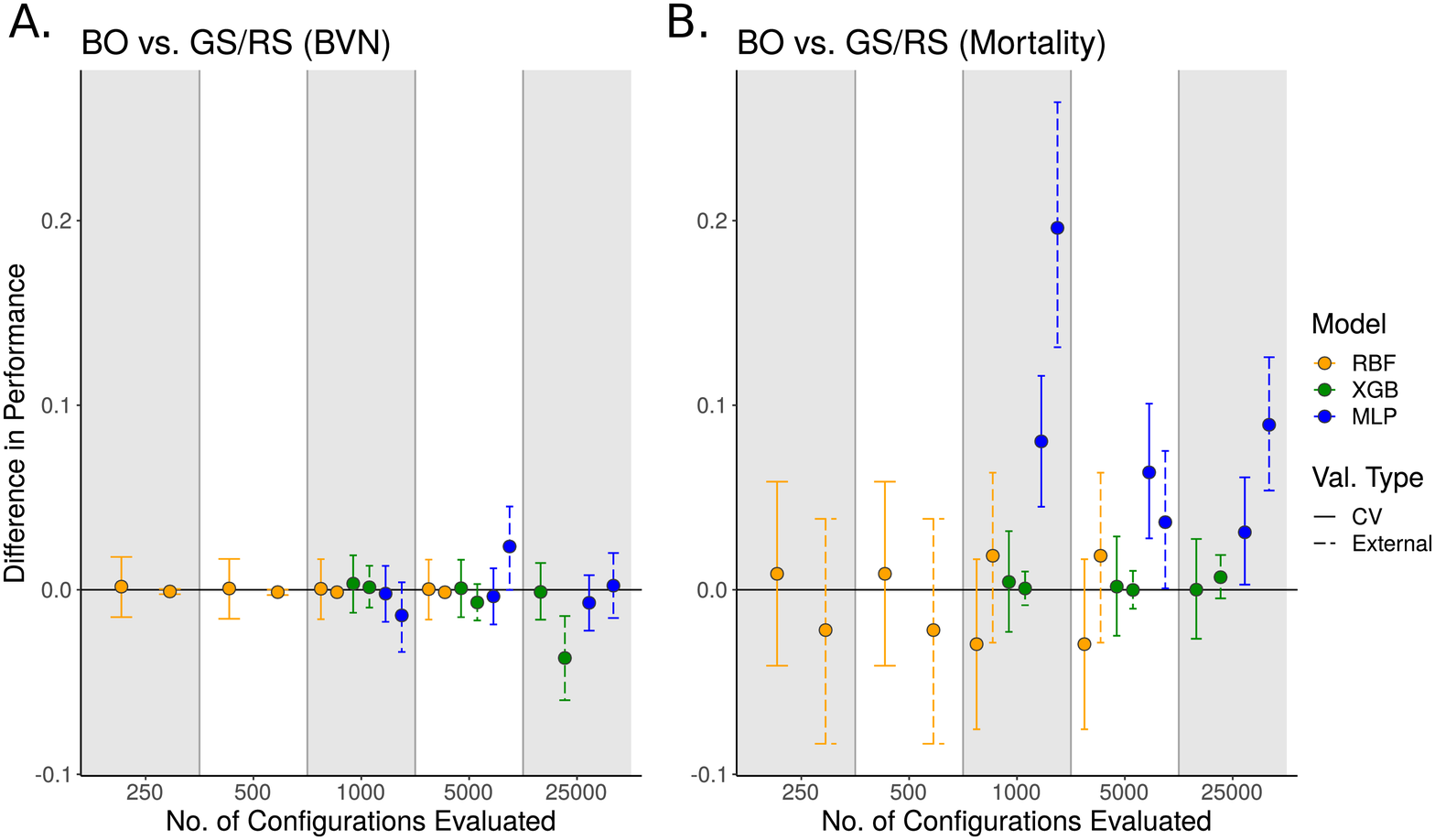}
\end{minipage}
\caption{Differences in classifier performance in internal (CV) and external validation on the BVN (A; mAUC) and mortality (B; AUC) tasks for GS/RS evaluation budgets with the expected improvement acquisition function and 25 initialization configurations in the \emph{transformed} hyperparameter space. Automatic relevance determination was \emph{enabled} for BO runs. Same run settings and figure layout as in Figure~\ref{fig:bo_ei_unit1_ard1_init25_direct_compare} except that the indicated evaluation budgets only apply to GS/RS-selected classifiers. BO-selected classifiers are taken from 100-evaluation (RBF) or 500-evaluation (XGB and MLP) runs.}
\label{fig:bo_ei_unit1_ard1_init25_long_range_compare}
\end{figure}

\begin{figure}[htbp]
\begin{minipage}{\textwidth}
  \centering
  \includegraphics[width=\textwidth]{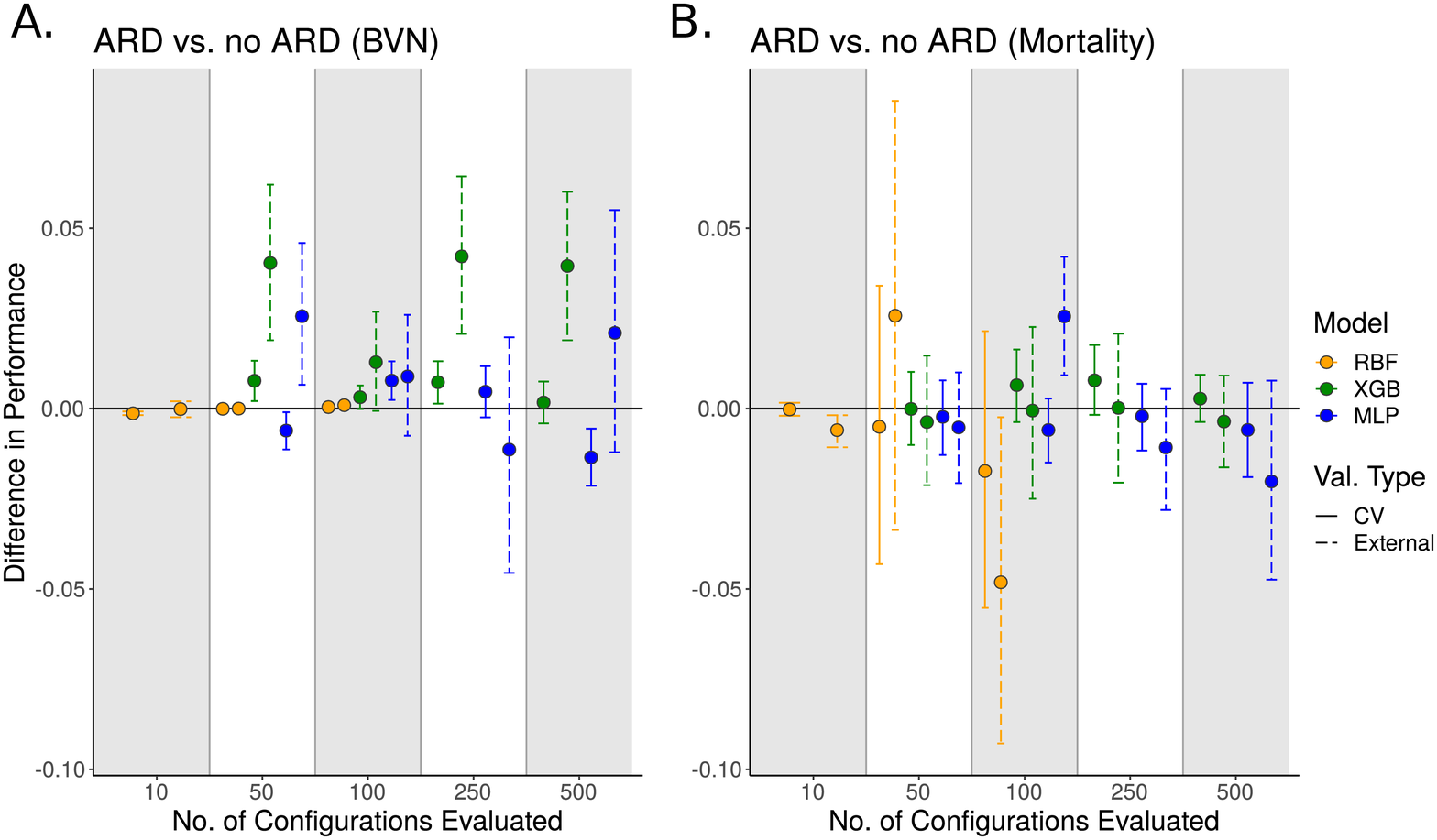}  
\end{minipage}
\caption{Differences in classifier performance in internal (CV) and external validation on the BVN (A; mAUC) and mortality (B; AUC) tasks for BO-selected classifiers with or without automatic relevance determination enabled. BO runs were performed with the upper confidence bound acquisition function and 25 initialization configurations in the \emph{original} hyperparameter space. Differences greater than 0 indicate better performance for the classifier selected by BO with ARD enabled. Points represent observed differences while error bars represent 95\% bootstrap confidence intervals.}
\label{fig:bo_ucb_ard_compare}
\end{figure}

\begin{figure}[htbp]
\begin{minipage}{\textwidth}
  \centering
  \includegraphics[width=\textwidth]{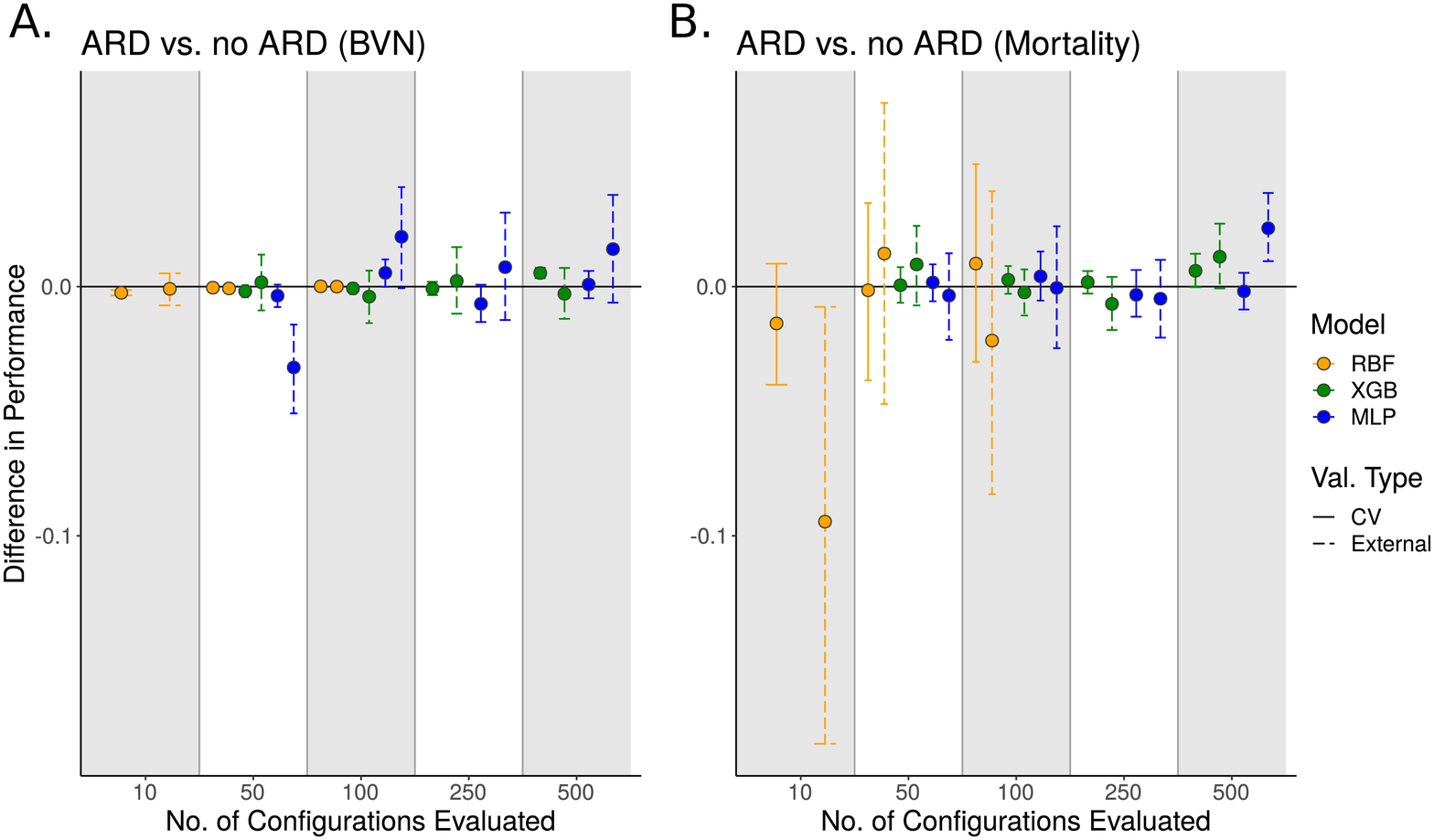}  
\end{minipage}
\caption{Differences in classifier performance in internal (CV) and external validation on the BVN (A; mAUC) and mortality (B; AUC) tasks for BO-selected classifiers with or without automatic relevance determination enabled. BO runs were performed with the expected improvement acquisition function and 25 initialization configurations in the \emph{transformed} hyperparameter space. Differences greater than 0 indicate better performance for the classifier selected by BO with ARD enabled. Points represent observed differences while error bars represent 95\% bootstrap confidence intervals.}
\label{fig:bo_unit1_ard_compare}
\end{figure}

\pagebreak

\bibliographystyle{plain}
\bibliography{supplement}